\documentclass[%
 reprint,
 amsmath,amssymb,
 aps,
 ]{revtex4-2}

 \usepackage{graphicx}% Include figure files
\usepackage{dcolumn}% Align table columns on decimal point
\usepackage{bm}% bold math
\usepackage{hyperref}% add hypertext capabilities
\usepackage{dsfont}
\newcommand{\vect}[1]{\boldsymbol{#1}}
\usepackage{amsmath}
\usepackage[vlined,ruled]{algorithm2e}
\usepackage{subfigure}
\usepackage[textsize=tiny]{todonotes}

\newcommand{\bs}[1]{\boldsymbol{#1}}

\newcommand{\MM}{$\mathcal{M}$}
\newcommand{\DD}{$\mathcal{D}$}
\usepackage{xcolor}
\newcommand{\params}{\bs{\theta}}

\begin{document}

\title{Comparing analytic and data-driven approaches to parameter identifiability:  \\ A power systems case study}%

\author{Nikolaos Evangelou}
\affiliation{Department of Chemical and Biomolecular Engineering, Johns Hopkins University, 3400 North Charles Street, Baltimore, MD 21218, USA}

\author{Alexander M. Stankovic}
\affiliation{Department of Electrical and Computer Engineering, Santa Clara University, 500 El Camino Real, Santa Clara, CA 95053, USA}

\author{Ioannis G. Kevrekidis}
\affiliation{Department of Chemical and Biomolecular Engineering, Johns Hopkins University, 3400 North Charles Street, Baltimore, MD 21218, USA}
\affiliation{Department of Applied Mathematics and Statistics, Johns Hopkins University, 3400 North Charles Street, Baltimore, MD 21218, USA}
\email{yannisk@jhu.edu}

\author{Mark K.~Transtrum}
\affiliation{Department of Physics and Astronomy, Brigham Young University, Provo UT 84602}
\email{mktranstrum@byu.edu}

\begin{abstract}
\noindent
  Parameter identifiability refers to the capability of accurately inferring the parameter values of a model from its observations (data). %and is a fundamental analysis when modeling data.
  Traditional analysis methods exploit analytical properties of the closed form model, in particular sensitivity analysis, to quantify the response of the model predictions to variations in parameters.
  Techniques developed to analyze data, specifically manifold learning methods, have the potential to complement, and even extend the scope of the traditional analytical approaches.
  We report on a study comparing and contrasting analytical and data-driven approaches to quantify parameter identifiability and, importantly, perform parameter reduction tasks.
  We use the infinite bus synchronous generator model, a well-understood model from the power systems domain, as our benchmark problem.
  Our traditional analysis methods use the Fisher Information Matrix to quantify parameter identifiability analysis, and the Manifold Boundary Approximation Method to perform parameter reduction.
  We compare these results to those arrived at through data-driven manifold learning schemes: Output - Diffusion Maps and Geometric Harmonics.
  For our test case, we find that the two suites of tools (analytical when a model is explicitly available, as well as data-driven when the model is lacking and only measurement data are available) give (correct) comparable results; these results are also in agreement with traditional analysis based on singular perturbation theory.
  We then discuss the prospects of using data-driven methods for such model analysis.
\end{abstract}

\maketitle

\section{Introduction}
\label{sec:introduction}

% Motivation for Parameter identifiability 
Recent decades have seen dramatic advances in the development and application of mathematical models for science and engineering applications.
Emerging technologies, such as digital twins\cite{juarez2021digital,huxoll2021model,thelen2022comprehens,fabiani2024task}, leverage large, high-fidelity simulations that blend physics- and data-driven approaches at unprecedented scales\cite{karpatne2017theory,willard2020integrating,rai2020driven,raissi2019physics,karniadakis2021physics}.
To be useful, these large simulations must provide quality fits to data, rigorous performance guarantees, and uncertainty quantification.
Traditional model analysis techniques leverage analytical properties of the model, such as sensitivity analysis, often in conjunction with expert intuition, and are difficult to apply to large models or to black-box, ``legacy" simulation algorithms.
However, many techniques for analyzing data also lend themselves to analyzing the underlying models, and have the potential to complement and extend traditional analytical tools\cite{holiday2019manifold,evangelou2022parameter}.
In this paper, we report on a parameter identifiability\cite{cobelli1980parameter,raue2014comparison,holiday2019manifold,evangelou2022parameter} and reduction\cite{zhan2017survey} study using both traditional, analytical methods and manifold learning techniques developed in the data science context.
We use a transient stability model from the power systems domain as a benchmark example to compare and contrast methods.
We quantify the general agreement between the methods, which is also consistent with the physics of synchronous generator dynamics.

Parameter identifiability refers to the ability of uniquely inferring a model's parameter values from data.
In the most abstract sense, a multi-parameter model is a mathematical mapping that takes parameters as inputs and makes predictions for quantities of interest.
At its core, the question of identifiability is whether this mapping from parameters to behaviors is injective, so that an inverse map can be defined.
If this is the case, then parameters can be uniquely learned (estimated) from data, and the model is said to be {\em identifiable}.
Otherwise, the model is said to be {\em structurally unidentifiable}.
In the analytical approach, local injectivity is assessed through the linearization of the model map.
If the matrix of the local linearization has full column rank, then a local neighborhood of the parameter space maps uniquely to behaviors and can be unambiguously identified from data.
Restated in the language of statistics, if the Fisher Information Matrix (FIM) is full rank, then the parameters are locally structurally identifiable\cite{rothenberg1971identificat,brouwer2018underlying}.
The nuances of fitting models to real-world, noisy data introduce further complications.
While a model may be structurally identifiable, noisy data may lead to unreasonably large parametric uncertainties.
This effect, known as \textit{practical unidentifiability}, has no consensus definition; it is, however, generally associated with an ill-conditioned FIM.

In the early 2000s, a systematic study of the Fisher Information Matrices of large, multi-parameter differential  equation system models from diverse fields demonstrated that they frequently exhibit common spectral features, known as {\em sloppiness}\cite{brown2004statistical,gutenkunst2007universally,transtrum2015perspective,quinn2022information}.
Universally, the information spectra of multi-parameter models are observed to exhibit eigenvalues that are nearly evenly spaced on a log scale over many orders of magnitude.
Colloquially, sloppy models are {\em practically unidentifiable}, with a particular hierarchy of relevant parameters.
Sloppiness has been observed in systems biology\cite{gutenkunst2007universally}, interatomic potentials\cite{frederiksen2004bayesian}, complex engineered systems\cite{waterfall2006sloppy,transtrum2018information}, critical phenomena\cite{machta2013parameter}, and others\cite{quinn2022information}.
Quantifiers of sloppiness have been extended beyond local, Fisher Information analyses, using Information Geometry\cite{transtrum2010why, transtrum2011geometry, quinn2019visualizing, quinn2019chebyshev}.

While there are many challenges associated with analyzing sloppy models, it was pointed out early on that practical science requires that systems be intrinsically unidentifiable in some respects\cite{machta2013parameter}.
Even the most detailed mathematical models omit irrelevant nuances of the physical system\cite{mattingly2018maximizing}, and modelers are quick to recite maxims such as ``All models are wrong''\cite{box1976science} or ``Don't model bulldozers in terms of quarks''\cite{goldenfeld1999simple}.
If a model's accuracy were conditional on its faithful representation of a physical system, only first-principles models built from a reductionist perspective would ever be useful.
Mathematical studies into the origin of sloppiness thus provide clues into how to understand the relationship between abstract mathematical representation and real, physical systems.
These questions are especially relevant given the growing integration of AI into the scientific process\cite{raissi2019physics,karniadakis2021physics,gunning2019xai} and digital twins\cite{juarez2021digital,fabiani2024task} built on top of high-fidelity simulations.

Because the range of behaviors of unidentifiable models has an effective dimensionality less than (often much less than) the explicit number of parameters, they can be difficult to analyze and interpret.
Key questions about how model behaviors relate to parameter values are obscured by the fact that identifiable/unidentifiable parameters are (often nonlinear) combinations of ``bare'' parameters in terms of which the model is formulated. 
High-dimensional parameter spaces exacerbate these technical challenges.
Parameter reduction methods often follow identifiability analysis, so as to parsimoniously characterize the system's parameter dependence and facilitate reasoning about the relationship between behaviors and parameters.
The Manifold Boundary Approximation Method (MBAM) is one such parameter reduction method that arose within the sloppiness literature\cite{transtrum2014model,transtrum2016bridging,transtrum2017measurement}.
Others such techniques include active subspaces\cite{constantine2015active} and soft set theory\cite{zhan2017survey}, 

Beyond such analytic approaches, however, manifold learning methods are particularly well-positioned for identifiability analysis and parameter reduction.
Previous work has developed conformal autoencoders\cite{peterfreund2020local} and Output - Diffusion Maps\cite{holiday2019manifold,evangelou2022parameter} for these purposes.
By directly sampling model behaviors, manifold learning can characterize the dimensionality of the response space (i.e., the number of identifiable parameter combinations) and learn the (potentially complicated) relationship between identifiable parameter combinations and the ``bare,'' physical parameters.

In this work, we advance the theory and practice of parameter identifiability and reduction by comparing traditional analytic approaches with data-driven, manifold learning techniques for dynamical models of transient disturbances in power systems.
Power systems are an ideal use case for this study because they are inherently multi-scale and multi-physics and known to exhibit sloppiness at different scales \cite{transtrum2018information}.
Furthermore, the scale of energy conversion and transmission systems make exhaustive modeling impractical, and modelers must constantly make informed tradeoffs between abstraction and fidelity depending on the target application.
These challenges are highlighted by the failure of detailed models to capture details in well-instrumented events, such as the 2003 blackout\cite{andersson2005causes}.
Thus, power systems exemplify a domain in which advances in parameter identifiability and reduction could potentially produce important advances in engineering design and in operation safety and reliability.

This paper is organized as follows:
In section~\ref{sec:power-systems-model} we introduce the details of our benchmark model.
Section \ref{sec:analytic-methods} review details of analytic approaches (Fisher Information analysis, information geometry, and MBAM in subsections~\ref{sec:FIM}, \ref{sec:infogeometry}, and \ref{sec:MBAM} respectively).
Section \ref{sec:data-methods} presents data-driven manifold learning methods (Output - Diffusion Maps, geometric harmonics, and the relevance of the implicit function theorem in subsections \ref{sec:OutputInformedDiffusionMaps}, \ref{sec:geometricharmonics}, and \ref{sec:inversefunctiontheorem} respectively).
Our results are presented in section~\ref{sec:results}.
In section~\ref{sec:discussion} we interpret the results in the context of power systems and discuss the outlook for data-driven techniques for parameter identifiability analysis and reduction.
Our primary results are to demonstrate that the two classes of methods agree with each other and give results consistent with the physical interpretation of synchronous generators in terms of singular perturbation analysis.
This motivates future applications of Output - Diffusion Maps and geometric harmonics to other systems that have proven less tractable to analytical techniques.

\section{Power Systems Model}
\label{sec:power-systems-model}

Power systems models are generally written in the differential algebraic equation (DAE)-based form:
\begin{align}
  \label{eq:DAEs}
  \dot{\bs{x}} & = \bs{f}( \bs{x}, \bs{z}, \params, t) \\
  0 & = \bs{g}( \bs{x}, \bs{z}, \params, t)
\end{align}
where $\bs{x}$ are the differential state variables, $\bs{z}$ are the algebraic variables, $\params$ are the parameters, and $t$ is the (scalar) time variable.
The system measurements are of the form
\begin{align}
  \bs{y} & = \bs{h}(\bs{x}, \bs{z}, \params, t)
\end{align}
where $\bs{h}$ denotes the observation function.
In this study, we use the so-called ``infinite bus'' generator model, described in detail in \cite{sauer2017power,transtrum2017measurement}, whose dynamic equations are
\begin{align}
  \label{eq:model_dyn}
  \dot{\delta} & = \omega_b (\omega - \omega_0) \\
  \dot{\omega} & = \frac{1}{H} \left( P_m - P_g - D (\omega - \omega_0)\right) \\
  \dot{e}'_q & = \frac{1}{T'_{d0}} \left( -e'_q - (x_d - x'_d) i_d + v_{f0})\right) \\
  \dot{e}'_d & = \frac{1}{T'_{q0}} \left( -e'_d + (x_q - x'_q) i_q\right) \\
  \dot{e}''_q & = \frac{1}{T''_{d0}} \left( -e''_q + e'_q - (x'_d - x''_d) i_d \right) \\
  \dot{e}''_d & = \frac{1}{T''_{q0}} \left( -e''_d + e'_d +  (x'_q - x''_q) i_q\right)  
\end{align}
and the algebraic equations are
\begin{align}
  \label{eq:model_alg}
  v_d & = V \sin (\delta - \vartheta) \\
  v_q & = V \cos (\delta - \vartheta) \\
  i_d & = (e''_q - v_q)/x''_d \\
  i_q & = (v_d - e''_q)/x''_q  \\
  P_g & = v_d i_d + v_q i_q.
\end{align}
\noindent
Note that $\delta$ represents the rotor angle, $\vartheta$ is the bus voltage angle, and $\omega_0$ is the reference angular velocity ($2\pi 60$ Hz).
Initial conditions, constants, and tunable parameters are given in Tables~\ref{tab:ics},~\ref{tab:constants}, and~\ref{tab:parameters} respectively.
These initial conditions are characteristic of a system state immediately following a dynamic disturbance (e.g., a short, a load change, etc.) and are not at steady state.
Assuming full state observation, we explore the dependence of the trajectory on the parameters for %both short,
%$t \in (0,3)$, and 
long times, here $t \in (3,5)$. This choice is motivated by physical considerations of power systems.
Real systems are characterized by a wide range of time scales, but models typically approximate the fastest time scales as algebraic equations through singular perturbation.
Transient stability studies typically ignore the fastest dynamics, in the spirit of singular perturbation analysis.
Such models are most faithful to the real system on longer time scales as those we have selected to study here.
These time scales are usually observed in practice and also constitute the dynamics targeted for control in power applications.

\begin{table}
  \centering
  \begin{tabular}{|c|c|c|}
    \hline
    Index & Name & Value \\
    \hline
    1 & $\delta(0)$ &  0.5 \\
    2 & $\omega(0)$ &  0.98 \\
    3 & $e'_{q}(0)$ &  2.13 \\
    4 & $e'_{d}(0)$ &  0.02 \\
    5 & $e''_{q}(0)$ &  1.93 \\
    6 & $e''_{d}(0)$ &  0.02 \\
    \hline
  \end{tabular}
  \caption{Initial Conditions}
  \label{tab:ics}
\end{table}

\begin{table}
  \centering
  \begin{tabular}{|c|c|c|}
    \hline
   Index & Name & Value \\
    \hline
   1 & $\omega_b$ & 120 $\pi$ \\
   2 & $v_{f0}$ & 4.2 \\
   3 & $P_m$ & 0.7 \\
   4 &  $V$ & 1.09 \\
   5 &  $\theta$ &  0 \\
    \hline               
  \end{tabular}
  \caption{Constant model parameters}
  \label{tab:constants}
\end{table}

\begin{table}
  \centering
  \begin{tabular}{|c|c|c|}
    \hline
    Index & Name & Value \\
    \hline
    1 & $H$ &  2.53 \\
    2 & $D$ &  0.5 \\
    3 & $x_d$ &  5 \\
    4 & $x_q$ &  4.88 \\
    5 & $x'_q$ &  2.86 \\
    6 & $x'_d$ &  0.928 \\
    7 & $x''_q$ &  0.48 \\
    8 & $x''_d$ &  0.48 \\
    9 & $T'_{d0}$ &  4.75 \\
    10 & $T''_{d0}$ &  0.06 \\
    11 & $T'_{q0}$ &  1.5 \\
    12 & $T''_{q0}$ &  0.21 \\
    \hline
  \end{tabular}
  \caption{Tunable  model parameters}
  \label{tab:parameters}
\end{table}

Physical considerations require that the parameters satisfy several constraints,
\begin{align}
  \label{eq:model_constraints}
  x_d & \geq x_q \geq x'_q \geq x'_d \geq x''_q \geq x''_d \geq 0 \\
  T'_{d0} & \geq T''_{d0} \geq 0 \\
  T'_{q0} & \geq T''_{q0} \geq 0, 
\end{align}
suggesting that they cannot be varied independently.
We therefore introduce a set of ``independently variable'' parameters in Table~\ref{tab:independentparameters}.
These parameters must be non-negative but can otherwise be varied independently.
We computationally explore the parameter dependence of the model by numerically solving the model equations as these parameters are varied by 10\% from their nominal values.
Because $\delta x_5 = 0$ already saturates its physical bound, we omitted this parameter from the analysis, leaving an 11-parameter model.

\begin{table}
  \centering
  \begin{tabular}{|c|c|c|}
    \hline
    Name & Definition & Value \\
    \hline
    $H$ & & 2.53 \\
    $D$ & &  0.5 \\
    $\delta x_1$ &  $x_d - x_q$ & 0.12 \\
    $\delta x_2$ & $x_q - x'_q$ & 2.02 \\
    $\delta x_3$ & $x'_q - x'_d$ & 1.932 \\
    $\delta x_4$ & $x'_d - x''_q$ & 0.448 \\
    $x''_d$ & &  0.48 \\
    $\delta T_d$ & $T'_{d0} - T''_{d0}$ & 4.69 \\
    $\delta T_q$ & $T'_{q0} - T''_{q0}$ & 1.29 \\
    $T''_{d0}$ & & 0.06 \\
    $T''_{q0}$ & & 0.21 \\
    $\delta x_5$ & $x''_q - x''_d$ & 0 \\
    \hline
  \end{tabular}
  \caption{Independent Parameters}
  \label{tab:independentparameters}
\end{table}

This model has the advantage of being small enough to enable extensive numerical exploration while still exhibiting the key properties of sloppiness.
Power systems models such as this are almost always fit to data by nonlinear least squares regression.
The statistical assumptions underlying this is that data are given by the model plus additive measurement noise,
\begin{align}
  \label{eq:datamodel}
  \bs{d}_i & = \bs{y}_{\params} (t_i) + \bs{\xi}_i
\end{align}
where $\bs{\xi}$ are i.i.d. zero-mean Gaussian random variables with variance $\sigma^2$.

It is convenient to introduce the data vector
\begin{align}
  \label{eq:datavec}
  \bs{d} & =
           \begin{pmatrix}
             \bs{d}_1 \\
             \bs{d}_2 \\
             \vdots \\
             \bs{d}_n 
           \end{pmatrix}
\end{align}
and the \emph{model map},
\begin{align}
  \label{eq:modelmap}
  \bs{Y}({\params}) & =
                   \begin{pmatrix}
                     \bs{y}_{\params} (t_1) \\
                     \bs{y}_{\params} (t_2) \\
                     \vdots \\
                     \bs{y}_{\params} (t_n)
                   \end{pmatrix}
\end{align}
as the concatenation of all the data and  model predictions, respectively.
We define the parameter space $\Theta$ as the set of all allowed values of the parameters and data space \DD~as $\mathbb{R}^M$ where $M = \dim \bs{d}$.
With these definitions, the model map is formally a mapping between parameter space and data space: $\bs{Y}: \Theta \rightarrow \mathcal{D}$.

Note that while $\bs{y}_{\params}(t)$ and the model map $\bs{Y}({\params})$ are deterministic, the data vector $\bs{d}$ is not.
Indeed, the data vector is a multivariate random variable distributed as $\bs{d} \sim \mathcal{N}( \bs{Y}({\params}), \sigma^2 \mathds{1})$, where $\mathds{1}$ is the identity matrix.
This distinction can often be confusing on first encounter because the model as defined by the DAEs is deterministic it does not appear to be a stochastic problem.
However, the output of the deterministic model will generally not match the data perfectly.
Modeling the discrepancy between model and data as a random variable gives statistical significance to dissimilarities and likelihoods for different parameter values.
The hyper-parameter $\sigma$ corresponds to the scale of the disagreement between model and data, i.e., measurement noise.
Small values of $\sigma$ correspond to an accurate model.

\section{Analytic Methods}
\label{sec:analytic-methods}

\subsection{Fisher Information Methods}
\label{sec:FIM}
Nearly all identifiability studies begin with a local analysis based on the Fisher Information Matrix (FIM).
Assume a family of probability distributions for observations $\bs{y}$ parameterized by ${\params}$, $P_{\params}(\bs{y})$, the FIM is the negative expected Hessian of the log-likelihood
\begin{align}
  \label{eq:FIM}
  \mathcal{I}_{\mu\nu} & = -\left\langle \frac{\partial^2 \ell}{\partial \theta_\mu \partial \theta_\nu} \right\rangle,
\end{align}
where $\langle \cdot \rangle$ denotes expectation value which is taken over the distribution of $\bs{d}$.

For the model in Eq.~\eqref{eq:datamodel}, the FIM can be succinctly expressed in terms of partial derivatives of the model map
\begin{align}
  \mathcal{I}_{\mu\nu} & = \frac{1}{\sigma^2} \frac{\partial \bs{Y}}{\partial \theta_\mu} \cdot \frac{\partial \bs{Y}}{\partial \theta_\nu}
\end{align}
where $\sigma^2$ is the variance of measurement noise, $\xi$.
Sensitivites $\partial \bs{Y}/\partial {\params}$ are typically calculated using automatic differentiation\cite{revels2016forward}. 
Because we are primarily interested in the rank and conditioning of the FIM as well as ordering the \emph{relative} importance of different parameter combinations, it is common to factor our the scale of the measurement noise, $\sigma$, or equivalently set $\sigma = 1$.
We adopt this convention throughout.

If the FIM is rank-deficient when evaluated at parameter values ${\params}_0$, the model is structurally unidentifiable in a neighborhood of ${\params}_0$.
Varying the model parameters in the null space of the FIM leaves the output of the model map unchanged.
We say that such a model is \emph{locally structurally unidentifiable}, and the parameter combinations associated with the null space are said to be the \emph{unidentifiable parameters}.

Beyond considering the rank of the FIM, it is useful to additionally consider its entire eigenspectrum, referred to here as the \textit{information spectrum}.
Ill-conditioned FIMs are usually associated with practical unidentifiability.
The eigenspace associated with small eigenvalues corresponds to combinations of parameters to which the model is insensitive.
This association suggests referring to these parameter combinations as ``practically unidentifiable.''
In this work we graphically summarize this analysis by plotting the FIM eigenvalues on a log scale and the participation factors of the parameters in each eigenmode as a heatmap, \ref{fig:SG11FIM}.

\subsection{Sloppy Model Analysis and Information Geometry}
\label{sec:infogeometry}

Sloppy model analysis is the result of a systematic study of the identifiability properties of multi-parameter models from diverse fields.
It was empirically observed that the information spectrum is consistently ill-conditioned, with eigenvalues nearly uniformly spaced on a log-scale.
Condition numbers larger than $10^6$ are common for sloppy models with about 10 parameters, while larger models with more parameters may exhibit more extreme condition numbers.
Why sloppiness is an ubiquitous phenomenon remains an open question; however, the best rationalizations to date use information geometry arguments\cite{transtrum2010why,transtrum2015perspective,quinn2019chebyshev,quinn2022information}.

Define $\Theta$ as the space of allowed parameter values.
The model manifold \MM~is defined as the image of $\Theta$ under the model map:
\begin{align}
  \label{eq:MM}
  \mathcal{M} & = \{ \bs{Y}({\params}) \ | \ {\params} \in \Theta \}.
\end{align}
The model manifold is generically a smooth manifold of dimension equal to the number of structurally identifiable parameters.
The FIM is the Riemannian metric on the model manifold.

Geodesics are distance-minimizing curves on Riemannian manifolds.
Let ${\params}(\tau)$ denote a geodesic curve parameterized by $\tau$; such a curve is found by solving the geodesic equation
\begin{align}
  \label{eq:geodesic}
  \frac{d^2}{d \tau^2} \theta^\mu & = -\Gamma^\mu_{\alpha \beta} \frac{d\theta^\alpha}{d \tau} \frac{d\theta^\beta}{d \tau}.
\end{align}
In Eq.~\eqref{eq:geodesic}, the superscripts index components of the parameter vector ${\params}$ and $\Gamma$ are the Riemann connection coefficients given by
\begin{align}
  \label{eq:Gamma}
  \Gamma^\mu_{\alpha \beta} & = \frac{1}{\sigma^2} \sum_\nu \left( \mathcal{I}^{-1}\right)^{\mu \nu} \frac{\partial \bs{Y}}{\partial \theta^\nu} \cdot \frac{\partial^2 \bs{Y}}{\partial \theta^\alpha \partial \theta^\beta}
\end{align}
We follow the formulation for solving Eq.~\eqref{eq:geodesic} in\cite{transtrum2011geometry}.
The geodesic equation is second order, meaning its solution requires one to specify both an initial point in the parameter space as well as an initial direction.
Furthermore, $\tau$ is proportional to the length of the geodesic curve on the model manifold.
A typical analysis consists of plotting each component of the parameter vector versus, $\theta^\mu(\tau)$ versus $\tau$.
When the geodesic encounters a boundary, it manifests as a finite-time singularity in which one or more components of ${\params}$ diverge.

Geodesics allow one to ``measure'' the model manifold in directions specified by the initial condition, making them a global extension of the FIM analysis.
Using this approach, a key result is that sloppy model manifolds are bounded, with a hierarchy of widths.
Manifold widths are measured by numerically calculating geodesics on the model manifold in orthogonal directions\cite{transtrum2011geometry}.
The hierarchy of widths is often compared to a \textit{ribbon} that has a long direction, a narrower width, and an even narrower height.
Sloppy model manifolds are, therefore, hyper-ribbons: the high-dimensional generalization of this pattern.
Because most of the dimensions are so much thinner than the primary directions, the manifold is often said to exhibit a ``low effective dimensionality.''
The parameter directions aligned with the long axes of the manifold are said to be ``stiff'' while those associated with the short directions are said to be ``sloppy''.
When we construct reduced order models as described in the next section, we directly confirm that the model manifold has this hyper-ribbon structure.

\subsection{Manifold Boundary Approximation Method (MBAM)}
\label{sec:MBAM}

Because sloppy manifolds have a low effective dimensionality, it is natural to seek a low dimensional approximations by removing unidentifiable parameters.
The boundaries of sloppy models are useful approximations for this purpose.
The boundaries of sloppy model manifolds are similar to a smoothly distorted polyhedron: a hierarchy of faces, edges, corners, hyper-corners, etc.
Formally, this structure is a hierarchical cell complex\cite{transtrum2014information}.
Each face on the model manifold is associated with a physically interpretable approximation.
For example, in dynamical models with multiple time scales, there is a face associated with each singularly perturbed approximation.

\begin{figure}  \includegraphics[width=\columnwidth]{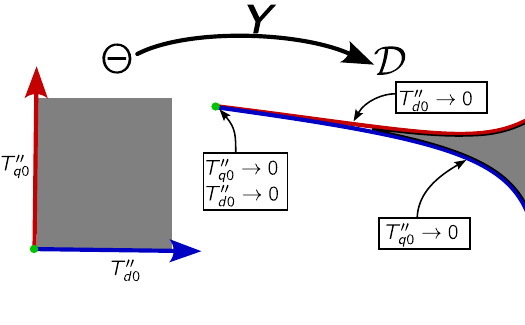}
  \caption{\textbf{The manifold boundary approximation method}.
    A model manifold, \MM, is the image of parameter space $\Theta$ in data space, \DD, under the model map $\bs{Y}$.
    Sloppy model manifolds are long and thin, exhibiting a low effective dimensionality.
    Here, the corner of the manifold is stretched out to be nearly one dimensional.
    Segments of the boundary correspond to different approximations, such as the two singularly perturbed limits  $T''_{d0} \rightarrow 0$ (red) and $T''_{q0} \rightarrow 0$ (blue).
    Since the model is sloppy, each of these segments capture most of the expressivity of the original model and either is a good approximation of the full manifold. 
  }
  \label{fig:mbam}
\end{figure}

This abstract concept is best understood graphically; consider the cartoon example in Figure~\ref{fig:mbam}.
The model manifold has the same dimensionality as the parameter space and in practice may be very high dimensional.
The parameter space and its image under the model map are represented by the gray, two-dimensional area in Figure~\ref{fig:mbam}.
Note, however, that the model manifold in data space is long and thin, suggesting that it may be approximated by a manifold of lower dimensionality.
Furthermore, the boundary of the manifold is divided into segments, each corresponding to a different approximation.
For our power system model, for example, there exists a singular limit corresponding to $T''_{d0} \rightarrow 0$ (red) and another corresponding to $T''_{q0} \rightarrow 0$ (blue).
Each of these singularly perturbed models is a boundary segment of co-dimension one.
They meet at a corner, i.e., a doubly singularly perturbed model of co-dimension two (green).
In the figure, either the red or blue segment capture most of the variation of the full model and represents a reasonable approximation.
In real-world models, the aspect ratios may be much more extreme so that low-dimensional approximations capture nearly all of the variation of the full model.

%%%YGK
% I suggest that there is an entire "blue ribbon" in parameter space close to the blue axis that maps to the blue boundary in the model manifold
% and a corresponding red ribbon -- maybe cross-hatched blue and red strips
% close to the axes on the left ? One suggestion is to change the arrows for Td''0 and Tq''0 that show the axis to be black and have two ribbons being blue and red showing the range of values in parameter space that correspond to the curves (blue and red) in the model manifold space.

The hierarchy of approximations represented graphically in Figure~\ref{fig:mbam} are the conceptual foundation of the Manifold Boundary Approximation Method (MBAM).
In this approach, we approximate the long, thin manifold by the section of its boundary aligned with the stiff direction\cite{transtrum2014model}.
For example, in the geometry of Figure~\ref{fig:mbam}, the model corresponding to one of the singularly perturbed approximations captures most of the expressive power of the original model because it is aligned with the long, stiff direction of the manifold.
These approximations are identified using geodesics.
A geodesic aligned with the short axis of the manifold will intersect a boundary segment aligned with the principal axis of the manifold.
The corresponding limit is inferred by inspecting the behavior of the parameters as the geodesic approaches the boundary.
Having found a boundary segment, the corresponding limiting approximation can be applied to the model equations to derive a new model with one less parameter.
Iterating this process removes the sloppy, practically unidentifiable parameters one at a time until a sufficiently parsimonious model is found.

\section{Manifold Learning Methods}
\label{sec:data-methods}

\subsection{Output - Diffusion Maps}
\label{sec:OutputInformedDiffusionMaps}
In this section, we begin by introducing the Diffusion Maps algorithm, illustrating its ability to reveal the intrinsic geometry of a data set $\mathbf{Y} = \{\vect{y}_i\}_{i=1}^N$ where $y_{i} \in \mathbf{R}^{m}$ sampled from a manifold $\mathcal{M} \in \mathbf{R}^{m}$. We then discuss the application of Diffusion Maps to the model output, referred to as \textit{Output Diffusion Maps}, that can be used for discovering the identifiable parameters of a (dynamical) system from samples\cite{evangelou2022parameter,holiday2019manifold}.

Given $\mathbf{Y}$, the Diffusion Maps algorithm first constructs an affinity matrix $\mathbf{A} \in \mathbb{R}^{N \times N}$. For the computation of $\mathbf{A}$ usually the Gaussian kernel,
\begin{equation}    \label{eq:affinity_kernel}
A(\vect{y}_i,\vect{y}_j) = \exp\bigg(\frac{- || \vect{y}_i - \vect{y}_j||_2^2}{2\varepsilon} \bigg),
\end{equation}
is used. In Equation \eqref{eq:affinity_kernel} $|| \cdot ||_2^2$ denotes the
$\ell^{2}$ norm squared that was used in our case; different metrics are also possible in practice. The parameter $\varepsilon >0$ controls the rate of the kernel's decay.

For the Diffusion Maps algorithm to discover the intrinsic geometry of the data regardless of the sampling density, the following normalization is applied \begin{equation}
    \mathbf{\Tilde{A}} = \mathbf{P}^{-1}\mathbf{A}\mathbf{P}^{-1},
\end{equation}
where $P_{ii}$ is a diagonal matrix defined as $P_{ii} = \sum_{j=1}^N A_{ij}$. The matrix $\mathbf{\tilde{A}}$ is normalized again to construct a row stochastic matrix $\mathbf{K}$,
\begin{equation}
    \mathbf{K} = \mathbf{D}^{-1}\mathbf{\tilde{A}}
\end{equation}
where $\mathbf{D}$ is defined as $D_{ii} = \sum_{j=1}^N \tilde{A}_{ij}$. The eigendecomposition of $\mathbf{K}$ is then computed that gives a set of eigenvectors $\vect{\phi}$ and eigenvalues $\lambda$
\begin{equation}
\mathbf{K}\vect{\phi}_i = \lambda_i \vect{\phi}_i.
\end{equation}

The eigenvectors, $\vect{\phi}$, provide an embedding of the original data set. However, some eigenvectors $\vect{\phi}$ parametrize the same eigendirections and they are therefore termed \textit{harmonic} eigenvectors. In practice, a proper selection of independent (non-harmonic) eigenvectors is necessary. This selection can be achieved by using the local-linear regression algorithm \cite{dsilva2018parsimonious}.
The number of non-harmonic eigenvectors $m^*$ is the effective dimensionality of the manifold, equal to the number of identifiable parameters.
If $m^*$ is smaller than the number of original dimensions, $m^* < m$, then dimensionality reduction is achieved. 

The Output - Diffusion Maps discovers the
identifiable parameters of a system from observations of its output.
More precisely, each data-point $\vect{y}_i \equiv Y({\params}_i)$
is a sample of the model map defined in Eq.~\eqref{eq:modelmap} for parameter vector ${\params}_i$.
The obtained non-harmonic eigenvectors in this case indicate the identifiable parameters of the system. Note that the identifiable parameters can be either some of the original parameters of the system or \textit{combinations} of the original parameters.

\subsection{Geometric Harmonics - Double DMaps}
\label{sec:geometricharmonics}
The Geometric Harmonics algorithm \cite{coifman2006geometric} is based on the Nystr\"om extension formula and initially has been proposed for extending a function $f$ sampled on the data set $\mathbf{Y}$ for out of sample data, $y_{new} \notin \mathbf{Y}$. For example, the identifiable parameters of a model could be thought as a smooth function $f$ on the model map $\bs{Y}({\params})$ (sampled behavior).
In this paper, we want to extend a function $f$, not in the original high dimensional coordinates $\bs{Y}({\params})$, but on the Diffusion Maps coordinates that provide an embedding of the intrinsic dimensionality of the system's behavior. This particular use case of Geometric Harmonics has been termed Double DMaps; and it has been presented in \cite{evangelou2023double} and successfully also applied in \cite{evangelou2022parameter,koronaki2023partial,koronaki2024nonlinear}. 

Similar to the Diffusion Maps algorithm, the first step in the computation here is to compute the affinity matrix $\mathbf{W} \in \mathbb{R}^{N \times N}$ in terms of a (Gaussian) kernel,

\begin{equation}    \label{eq:affinity_kernel_GH}
W(\vect{\phi}_i,\vect{\phi}_j) = \exp\bigg(\frac{- || \vect{\phi}_i - \vect{\phi}_j||_2^2}{2\varepsilon^{\star}} \bigg).
\end{equation}
Note that similarity distance in this case is computed using the discovered non-harmonic eigenvectors $\mathbf{\Phi} \in \mathbb{R}^{N \times m^{\star}}$.
The eigendecomposition of $\mathbf{W}$ gives a set of  
orthonormal eigenvectors  $\psi_0,\psi_1,\ldots,\psi_{N-1}$ and non-negative eigenvalues ($\sigma_0\geq\sigma_1\geq\cdots\geq\sigma_{N-1}\geq0$).  In practice, a subset of these eigenvectors, for $\delta>0$ $S_{\delta}=\{\alpha\,:\,\sigma_{\alpha}>\delta\sigma_{0}\}$ is computed to avoid numerical issues. 
This set of eigenvectors is used as a basis to project the function $f$. 

\begin{equation*}
    {f} \mapsto P_{\delta}{f} =  \sum_{\alpha \in S_{\delta}} \langle {f},\psi_{\alpha} \rangle \psi_{\alpha}, 
\end{equation*}

\noindent
where $\langle \cdot, \cdot \rangle$ denotes the inner product. The projection needs to be performed only once.
\noindent
At inference time, we would like to extend the function $f$ for out of sample points,  $\vect{\phi}_{new}\notin\mathbf{\Phi}$. This is achieved by implementing the Nystr\"om Extension formula,
\begin{equation}
\Psi_{\alpha}(\vect{\phi}_{new}) = \frac{1}{\sigma_{\alpha}} \sum_{j=1}^N \exp \bigg(\frac{-\| \phi_{new} - \phi_j \|^2_2}{2\epsilon^{\star}}\bigg)\vect{\psi}_{\alpha}(\vect{\phi}_j),
\label{eq:GH_extension}
\end{equation}
where $\vect{\psi}_{\alpha}(\phi_j)$ denotes the $j^{\text{th}}$ components of the eigenvector $\vect{\psi}_{\alpha}$ and $\Psi_{\alpha} (\vect{\phi}_{new})$ are the extended eigenvectors at $\vect{\phi}_{new}$.
The extension of $f$ is then obtained as
\begin{equation}
\label{eq:Geometric_Harmonics_Extension}
    (Ef)(\vect{\phi}_{new}) = \sum_{{\alpha\in{S_{\delta}}}} \langle f,\vect{\psi}_{\alpha}\rangle\vect{\Psi}_{\alpha}(\vect{\phi}_{new})\,.
\end{equation}  

\noindent
Apart from the extension of a function of interest defined on the data  $\mathbf{\Phi}$ (or $\mathbf{Y}$) Geometric Harmonics can be used to estimate a function's gradient
in terms of the input variables.
Symbolic differentiation of Equation \eqref{eq:Geometric_Harmonics_Extension} gives a closed form expression of the gradient of $f$. The expression of $\nabla f$ evaluated at $\vect{\phi}_{new}$ reads

\begin{multline}
    \label{eq:grad_GH}
    \nabla f (\vect{\phi}_{new}) = \\
    \sum_{\alpha \in S_{\delta}} \frac{\langle f, \vect{\psi}_{\alpha} \rangle}{\sigma_{\alpha}} \sum_{j=1}^N  \frac{( \vect{\phi}_j - \vect{\phi}_{new})}{\varepsilon^{\star}}\exp\bigg(\frac{ - \| \phi_{new} - \phi_j \|^2_2}{2\epsilon^{\star}}\bigg)\vect{\psi}_{\alpha}(\vect{\phi}_j).
\end{multline}

\subsection{Exploiting the Inverse Function Theorem}
\label{sec:inversefunctiontheorem}
In this section, we provide a brief description of our use of the Inverse Function Theorem (IFT). The IFT offers theoretical guarantees that a function $f$ is locally invertible \cite{marsden1993elementary}.

The IFT states that for a differentiable (vector) function \hbox{$f(\vect{x}) = \vect{y}$}, if the Jacobian matrix

\begin{equation}
    \label{eqn:jacobian}
    \mathbf{J}_f(\vect{x})=\begin{pmatrix}
    \frac{\partial f_1}{\partial x_1} & \cdots & \frac{\partial f_1}{\partial x_n}\\
        \vdots & \ddots & \vdots\\
        \frac{\partial f_n}{\partial x_1} & \cdots & \frac{\partial f_n}{\partial x_n}
    \end{pmatrix}
\end{equation}
\noindent
is invertible (has non-zero determinant), then $f$ behaves like a ``local'' bijection around $\vect{x}$.  Therefore, showing that the det$(\mathbf{J}_f(\vect{x}))$ has values that do not change sign,  guarantees that the mapping is locally invertible and thus one-to-one. 
\textit{In this work, we utilize the IFT to confirm that the Diffusion Maps coordinates, $\vect{\phi}$, are locally one-to-one with a set of physical parameters $\vect{{\params}}$ (and vice-versa).} In our case, the function $f$ is the regression function constructed with the Double DMaps - Geometric Harmonics. Please note that it is
possible a function $f$ to be locally invertible everywhere but not globally invertible. 

As a comprehensive measure of global invertibility, we report the mean absolute error of the map $f$, alongside parity plots that compare the true values with the predicted measurements.

%\textcolor{red}{check if/what we might want to say for global invertibility too.}

\section{Results}
\label{sec:results}

\subsection{The Infinite Bus Synchronous Generator}

\begin{figure}  \includegraphics[width=\columnwidth]{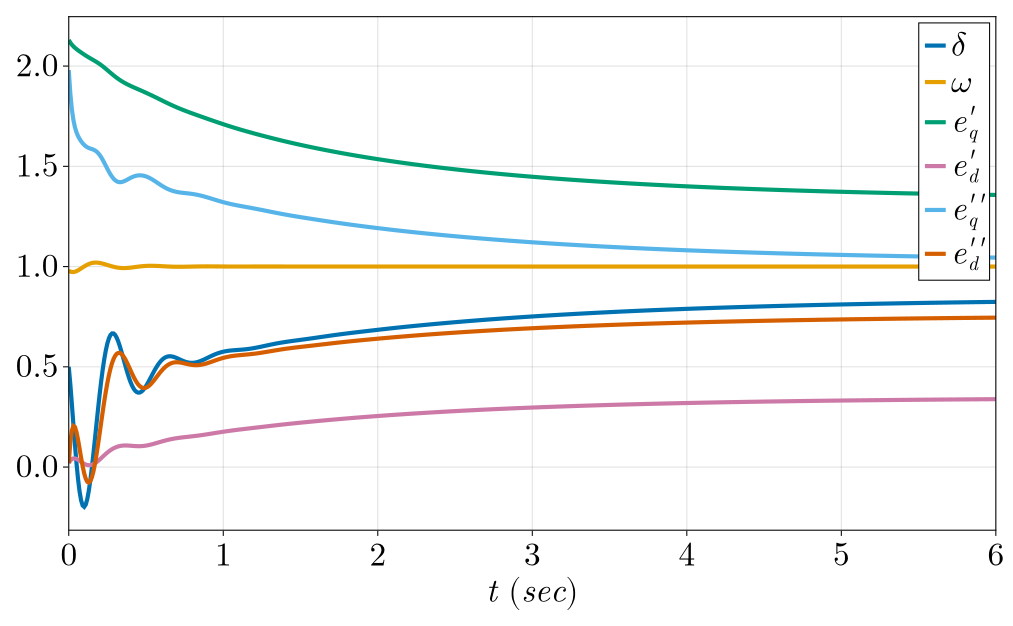}
  \caption{Dynamic Response of the model at the default parameter values.}
  \label{fig:dynamicresponse}
\end{figure}

Figure~\ref{fig:dynamicresponse} shows the dynamic response of the model for the nominal parameter values.
From a power systems transient stability perspective, the relevant question is to determine which parameters are identifiable from long-time dynamics.
Here, ``long times'' refers to a few seconds.
Therefore, in the following analysis we sample time points $t \in [3, 5]$ and ignore early parts of the transient.
From the figure we see that this region of the dynamics clearly captures the long-time decay toward equilibrium.

\subsubsection{MBAM}
\label{sec:infinitebusresults}

\begin{figure}
  \centering
  \includegraphics[width=\columnwidth]{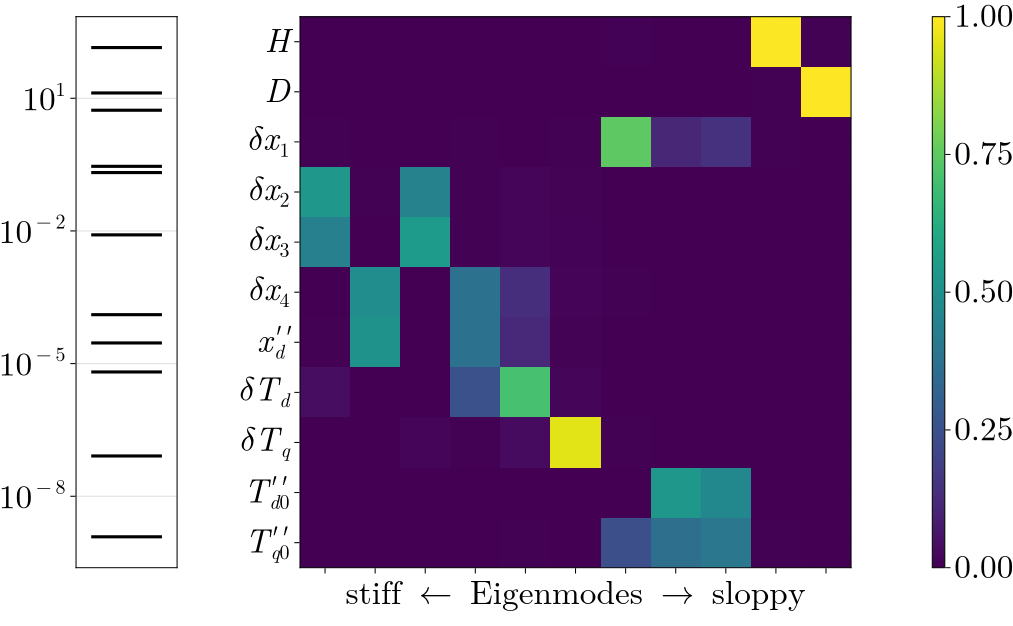}
  \caption{Information Spectrum of the Synchronous Generator model for long time dynamics; the participation of the ``bare'' parameters in the spectrum eigenvectors is given in the form of a heat map (color scale on the right).}
  \label{fig:SG11FIM}
\end{figure}

We first conduct a local identifiability analysis, summarized in Figure~\ref{fig:SG11FIM}.
Notice that the model is sloppy: the FIM eigenvalues span approximately 11 orders of magnitude and are nearly uniformly spaced over this range.
This is consistent with previous studies in power systems\cite{transtrum2017measurement,transtrum2018information}.
Because there is no clear gap in the eigenspectrum, it is difficult to attach a clear dimensionality to the manifold.
Since the model is evaluated in dimensionless units, it is natural to compare eigenvalues relative to one.
We take a cutoff at $10^{-2}$, corresponding to errors of a few percent, which suggests that the model is effectively six-dimensional at this resolution.
We will further justify this preliminary conclusion in subsequent analysis.

We now consider the participation factors of any single ``bare'' parameters in the eigenmodes.
The columns in Figure~\ref{fig:SG11FIM}, moving from left to right, are associated with eigenvalues in decreasing order.
We begin by considering the most identifiable parameter combinations, that is, the left most columns.
The first column is dominated by parameters $\delta x_2$ and $\delta x_3$, so the most identifiable parameter combination does not align with any bare parameter but is rather a mixture.
However, if we take the first six columns collectively, we see that they align neatly with exactly six bare parameters: $\delta x_2$, $\delta x_3$, $\delta x_4$, $x''_d$, $\delta T_d$, and $\delta T_q$.

The least identifiable parameter combination (right-most column) is nearly perfectly aligned with parameter $D$ and the next-least identifiable mode is aligned with parameter $H$.
The next several sloppy modes are combinations of parameters $\delta x_1$, $T'_{d0}$ and $T_{q0}$.
However, if we consider the five-dimensional subspace of the least identifiable parameters (i.e., the orthogonal complement to the identifiable subspace), it aligns with the five parameters $H$, $D$, $\delta x_1$, $T'_{d0}$ and $T_{q0}$.
The near perfect partitioning of the identifiable and unidentifiable subspaces into subsets of the bare parameters will be confirmed below by non-local analysis using both geodesics and Diffusion Maps.

\begin{figure}
  \centering
  \includegraphics[width=\columnwidth]{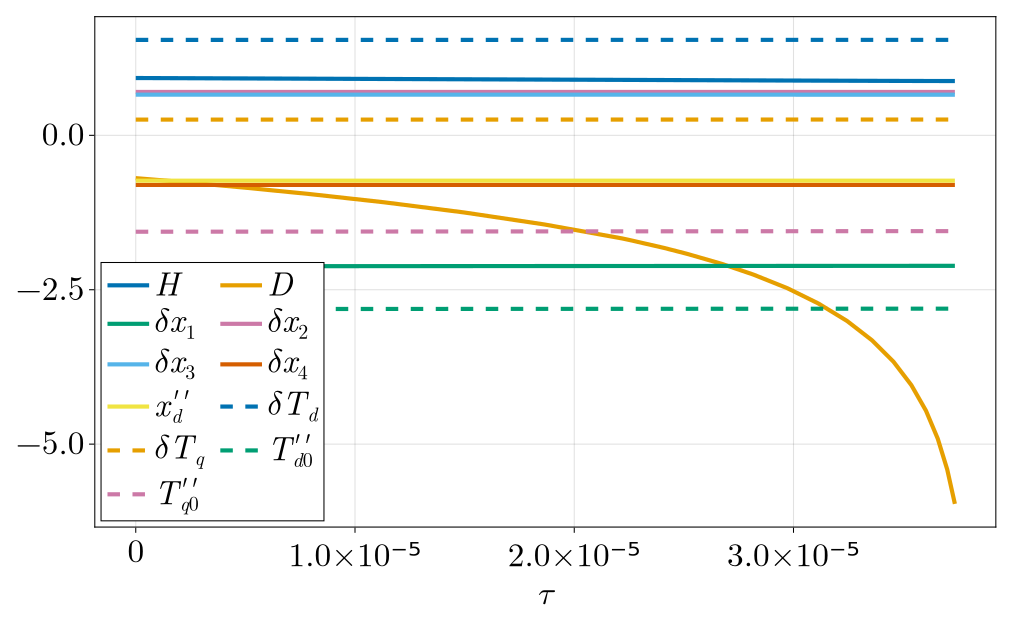}
  \caption{Geodesic oriented in sloppiest direction of the full model manifold in log-parameters.}
  \label{fig:geo11}
\end{figure}

We now nonlocally extend the sloppy model analysis using the Manifold Boundary Approximation Method.
Figure~\ref{fig:geo11} shows the numerical solution to the geodesic equation.
The horizontal axis, $\tau$ is the distance on the model manifold and each curve is the value of a parameter along the geodesic.
The geodesic originates from the nominal parameter values in the direction given by the least identifiable (i.e., most sloppy) parameter.
From Figure~\ref{fig:SG11FIM}, we see the initial direction is oriented almost exclusively in the direction of parameter $D$.
Moving along the geodesic, only parameter $D$ changes.

The curves in Figure \ref{fig:geo11} are on a log-scale, so that as the curve approaches a vertical asymptote around $\tau \rightarrow 3.8 \times 10^{-5}$, the parameter $D \rightarrow 0$.
The finite-time singularity of the geodesic indicates that it has reached a boundary after traveling a total distance of approximation $3.8 \times 10^{-5}$ units on the model manifold.
Previous work has indicated that the square root eigenvalues of the FIM are a useful estimate of the distance to a boundary\cite{transtrum2010why,transtrum2011geometry}.
The FIM eigenvalue associated with this direction is $(3.47\times 10^{-5})^2$, so this indeed provides a good approximation in this case.
Our conclusion from inspecting Figure~\ref{fig:geo11}, is that the nearest boundary is associated with the limit $D \rightarrow 0$.
Applying this limit leads to a new model with one fewer parameter.

Iterating this procedure removes the least identifiable parameters, one at a time, by projecting them along the thinnest axis of the model manifold onto an appropriate boundary approximation.
The limit $D\rightarrow 0$, took us from $11$ to $10$ parameters.
The next iteration traces a geodesic on the $10$ parameter model in the direction $H \rightarrow 0$ in similar fashion.
This limit is consistent with the local analysis in Figure~\ref{fig:SG11FIM}: notice the second sloppiest mode is dominated by the parameter $H$.
This second limit is a singular limit in which the generated power, $P_g = v_d i_d + v_q i_q$, is algebraically slaved to the mechanical power $P_m$.

\begin{figure}
  \centering
  \includegraphics[width=\columnwidth]{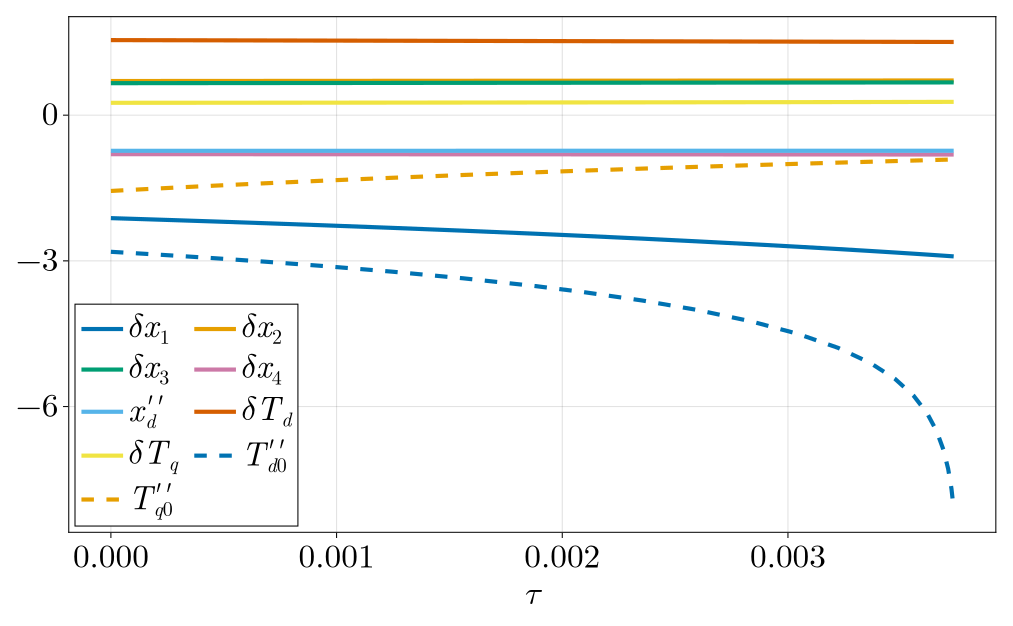}  
  \caption{Geodesic oriented in sloppiest direction of the 9 parameter manifold in log-parameters.}
  \label{fig:geo09}
\end{figure}

The third iteration is less obvious from the local analysis.
The third sloppiest mode in Figure~\ref{fig:geo11} is a mixture of several parameters.
The result of the geodesic is shown in Figure \ref{fig:geo09}.
The initial direction involves a mixture of parameters $T''_{q0}$, $T''_{d0}$, and $\delta x_1$, each of which has visibly shifted from their initial values by the end of the geodesic.
However, by the time it encounters the boundary, the geodesic curve rotates, so as to be dominated by the direction $T''_{d0} \rightarrow 0$.
We therefore conclude that the appropriate approximation is the associated singular limit.

In a similar manner, we remove additional parameters from the model.
The full sequence of approximations are summarized in Table~\ref{tab:mbam}.

\begin{table}
  \centering
  \begin{tabular}{|c|c|}
    \hline
    Reduction & Approximation \\
    \hline
    $11 \rightarrow 10$ parameters & $D \rightarrow 0$ \\
    $10 \rightarrow 9$ parameters & $H \rightarrow 0$ \\
    $9 \rightarrow 8$ parameters & $T''_{d0} \rightarrow 0$ \\
    $8 \rightarrow 7$ parameters & $T''_{q0} \rightarrow 0$ \\
    $7 \rightarrow 6$ parameters & $\delta x_1 \rightarrow 0$ \\
    \hline
  \end{tabular}
  \caption{Summary of MBAM limits}
  \label{tab:mbam}
\end{table}

The reduced model is  given by the equations
\begin{align}
  \label{eq:model_red}
  \dot{\omega} & = \frac{\ddot{\delta}}{\omega_b} \\
  \dot{e}'_q & = \frac{1}{T'_{d0}} \left( -e'_q - (x_d - x'_d) i_d + v_{f0})\right) \\
  \dot{e}'_d & = \frac{1}{T'_{q0}} \left( -e'_d + (x_d - x'_q) i_q\right) \\
  v_d & = V \sin (\delta - \theta) \\
  v_q & = V \cos (\delta - \theta) \\
  i_d & = (e''_q - v_q)/x''_d \\
  i_q & = (v_d - e''_q)/x''_q  \\
  P_m & = P_g = v_d i_d + v_q i_q \\
  e''_q & =  e'_q - (x'_d - x''_d) i_d \\
  e''_d & = e'_d +  (x'_q - x''_q) i_q.
\end{align}
The system is now third order (the states $\omega$, $e'_q$ and $e'_d$ are the remaining dynamic state variables).
The algebraic structure of this model is unusual and merits special explanation.
The state variable $\delta$ is an algebraic variable; it is defined implicitly by the equation $P_g = P_m$.
Having solved this equation, $\delta$, is a smooth function of time and can be numerically differentiated.
The variable $\omega$ is a dynamic variable whose time derivative is equal to the second time derivative of $\delta$, denoted by the double dot notation, $\ddot{\delta}$.

\begin{figure}
  \centering
  \includegraphics[width=\columnwidth]{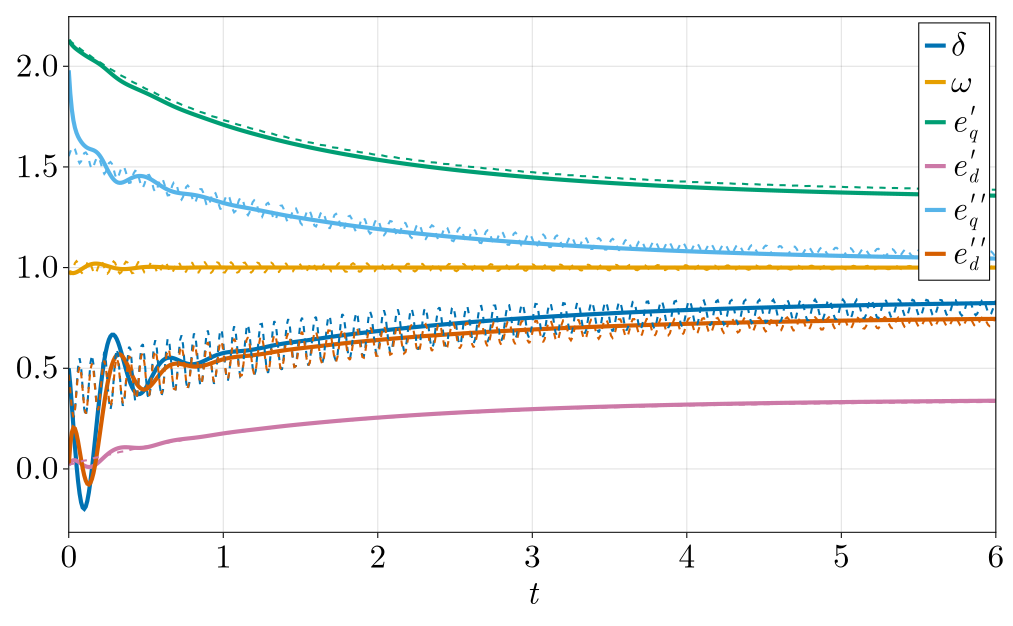}
  \caption{Full (solid) vs. Reduced (dashed) dyanmics}
  \label{fig:reduceddynamics}
\end{figure}

The dynamics of these equations, compared against the original dynamics, are shown in Figure \ref{fig:reduceddynamics}.
Notice that the errors are on average less than a few percent for the observations $\delta$, $\omega$, $e'_d$ and $e'_q$.
For the remaining state variables, $e''_d$ and $e''_q$, the reduced models has introduced high-frequency oscillations, characteristic of approximating these modes by the algebraic relations, which introduce errors about at most 10\%.

\begin{figure}
  \centering
  \includegraphics[width=\columnwidth]{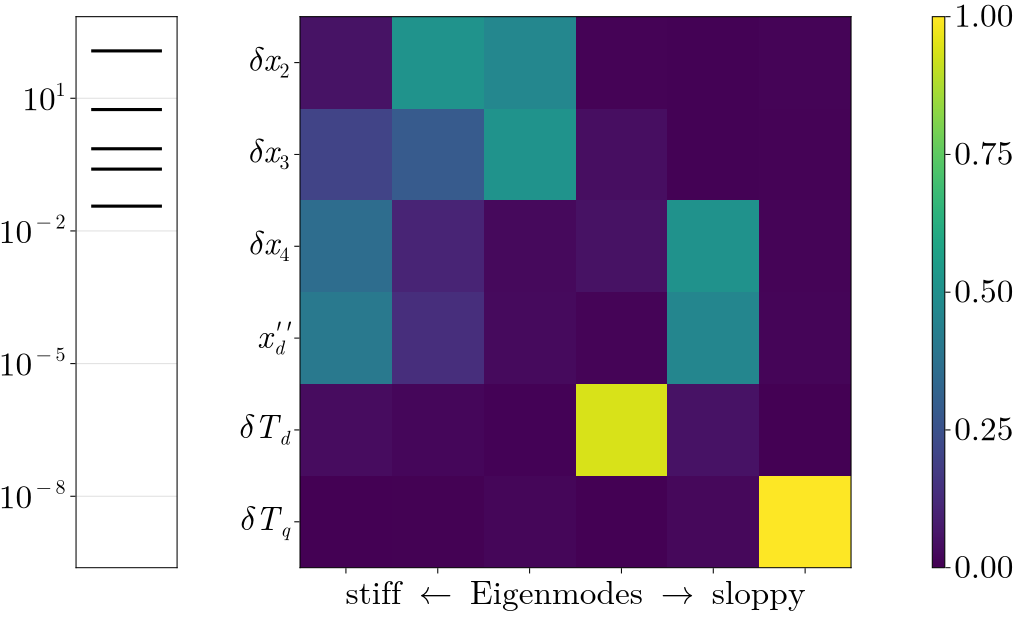}
  \caption{Information Spectrum for the reduced 6 parameter model, with participation of the the bare parameter components in the heat map on the right.}
  \label{fig:SG06FIM}
\end{figure}

The reduced model has only six parameters.
An analysis of the information spectrum of this model is shown in Figure~\ref{fig:SG06FIM}.
For reference, the scale of the eigenvalue plot is the same as that in Figure~\ref{fig:SG11FIM}.
The MBAM reduction has functionally removed the unidentifiable combinations, and the remaining six parameters have similar (though not exactly the same) participation factors in the identifiable subspace as the original model.

\subsection{Manifold Learning Computations}

\begin{table*}[ht!]
  \centering
  \begin{tabular}{|c|c|c|c|c|c|c|c|c|c|c|}
    \hline
    $x_{d}^{''}$ & $\delta x_{4}$ & $\delta x_{2}$ & $\delta x_{3}$ & $\delta T_q$ & $\delta T_{d}$ & $\delta {x}_1$ & $H$ & $T_{q0}^{''}$ & $D$ & $T_{d0}^{''}$     \\
    \hline
    0.0018 & 0.0026 & 0.0057 & 0.0072 & 
    0.0095& 0.0090 &
    0.2507 & 0.2532 &
    0.2534 & 0.2557 & 0.2534
    \\
    \hline
  \end{tabular}
  \caption{Mean absolute error values for the regressed parameters on the test set (rescaled between 0 and 1).
    }
  \label{tab:MAE_GH}
\end{table*}

\begin{figure}[htbp!]
    \centering
    \includegraphics[height = 6cm]{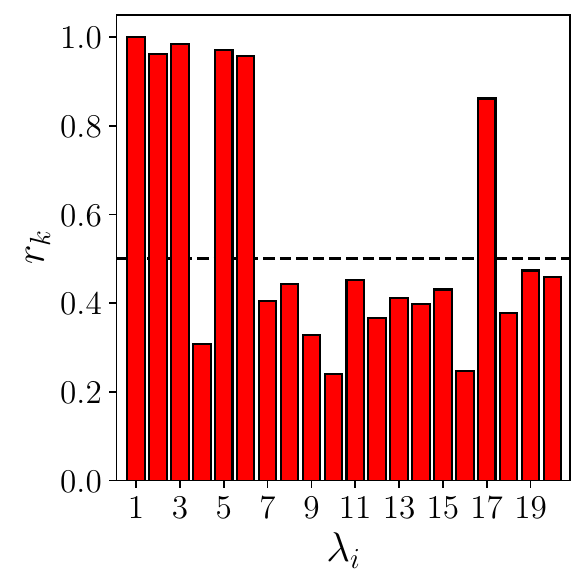}
    \label{fig:local_linear_regression}
    \caption{The residual ($r_k$) from the local linear regression algorithm suggesting that the output is six-dimensional with the eigenvectors $\vect{\phi} = \{\phi_1,\phi_2,\phi_3,\phi_5,\phi_6,\phi_{17}\}$}.
\end{figure}

For our analysis with the Output - Diffusion Maps, we started by fixing the initial condition to the value shown in Table \ref{tab:MAE_GH}. The parameter values shown in Table \ref{tab:MAE_GH} were chosen as the base parameter values.
 We then initially perturbed the values of the ``bare'' parameters by up to 10$\%$ from the base parameter values and integrated the ODE model dynamics for each perturbed value using the \textit{ode45} Matlab library until $T=5$, collecting the output every $dt=0.02$. The absolute and relative tolerance of the \textit{ode45} solver was set to $10^{-7}$. This process was subsequently repeated 10,000 times, and each time we collected the output for a different value of the independent parameters. 
For further analysis, we used the obtained trajectories after $T=3$ similar to the MBAM analysis.
For each distinct (perturbed) value of the parameters, we concatenated the outputs and rescaled the data between 0 and 1. This scaling step is not strictly necessary, but it is useful for improving the conditioning of the calculation.
The scale parameter, $\epsilon$, for Diffusion Maps was computed as a multiple of the median pairwise distances between the data points. The value of $\epsilon = 220.55$ was used for our experiments. 
To discover which are the non-harmonic Diffusion Maps coordinates the local-linear regression algorithm \cite{dsilva2018parsimonious} was used. We infer the dimensionality of the data by looking for an obvious gap in the residuals in Figure \ref{fig:local_linear_regression}.
By inspection, the local linear regression algorithm suggest that the response is six-dimensional, consistent with the FIM analysis above, and that the Diffusion Maps coordinates $\vect{\Phi} = \{\phi_1,\phi_2,\phi_3,\phi_5,\phi_6,\phi_{17} \}$ are non-harmonic. Since the response of the system is six dimensional, there are six identifiable parameters that affect the response of the system, and the remaining parameters are nonidentifiable.

Given these six data-driven coordinates, we want to check which of the independent model parameters are identifiable. To confirm this, we applied the Double-DMaps Geometric Harmonic algorithm to learn the model parameters as a function (a mapping) from the non-harmonic Diffusion Maps coordinates $\vect{\phi}$. For the regression scheme, we split the data into 80$\%$ for training and 20$\%$ for testing. We used a large number of eigenvectors (250 of them) as basis vectors, and the kernel parameter was estimated as the median of the pairwise distances, the data were rescaled to 0 and 1. 
\begin{figure}[htbp!]
    \centering
    \includegraphics[height = 6cm]{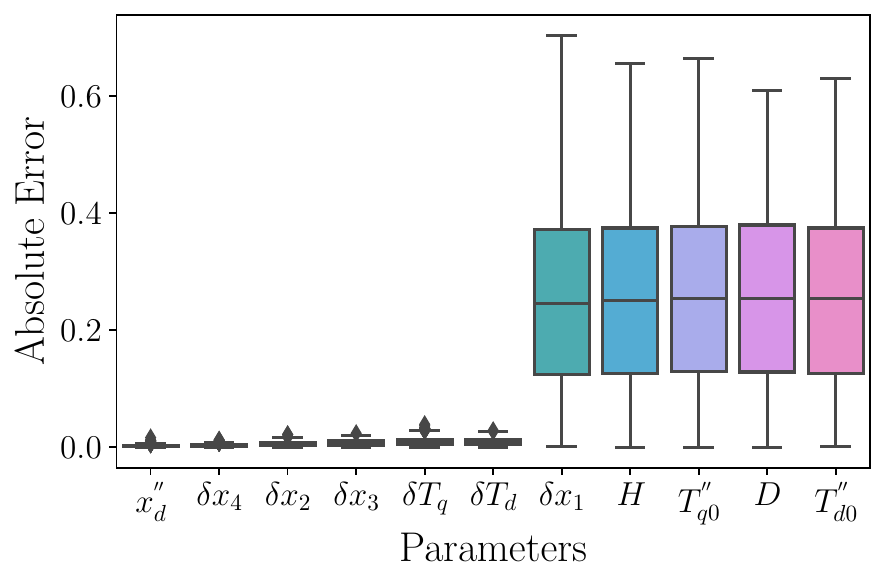}
\caption{The absolute (point-wise) error between the true original
parameters and the predicted with Double DMaps - Geometric Harmonic. The first six parameters with smallest absolute error are identifiable from the response.}
\label{fig:GHBoxParameters}
\end{figure}
In Figure \ref{fig:GHBoxParameters} we illustrate the absolute error between the true independent parameter values and the predictions with Double DMaps - Geometric Harmonics. The six independent parameters $\vect{p} = \{ x_{d}^{''}, \delta x_{4}, \delta x_{2}, \delta x_{3}, \delta T_q, \delta T_{d} \}$  were  approximated with an absolute error two order of magnitudes smaller than the remaining parameters (non-identifiable). This can be also seen from Table \ref{tab:MAE_GH}, where the mean absolute values of the regressed parameters are reported. For completeness we also provide the parity plots for training and test points in Figure \ref{fig:parity_plots_GH_parameters} of the Appendix.

Since six of the independent parameters can be well approximated as functions of the six Diffusion Maps coordinates  $\vect{\phi}$, we can claim that these six independent parameters $\vect{p}$ are identifiable.
We wanted also to check if the map $f:\vect{\phi} \mapsto \vect{p}$ is invertible and thus the two quantities are \textit{locally} one-to-one. To this end, we use the symbolic gradient of the Double-DMaps Geometric Harmonics (see Equation \eqref{eq:grad_GH}), apply the IFT and compute the determinant of the Jacobian for $f$ on the test set.

\begin{figure}[htbp!]
    \centering
    \includegraphics[height = 6cm]{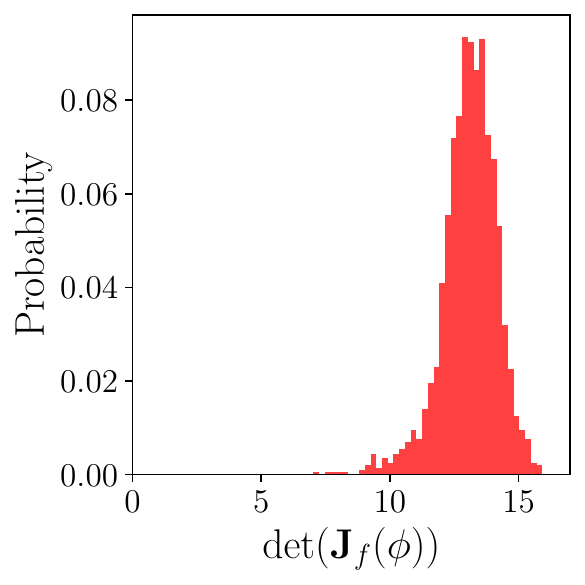}
    \caption{The histogram of the determinant of the Jacobian $\text{det}(\mathbf{J}f (\phi))$ computed on the test data with symbolic differentiation of the Double DMaps - Geometric Harmonics formula.}
\label{fig:JacobianPhis}
\end{figure}
As shown in Figure \ref{fig:JacobianPhis} the Jacobian is bounded away from zero (in this case, it is always positive) along our dataset, corroborating the invertibility of the $\vect{\phi}$ with respect to the $\vect{p}$.

\begin{figure}[htbp!]
    \centering
\includegraphics[height = 6cm]{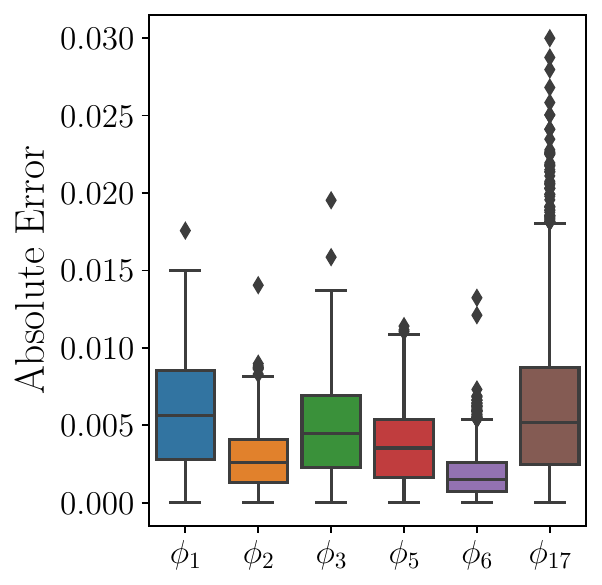}
    \caption{The absolute (point-wise) error between the true Diffusion Maps' coordinates values and the predicted with. 
Geometric Harmonic.}
    \label{GHBoxPhis}
\end{figure}

\begin{figure}[htbp!]
    \centering
    \includegraphics[height = 6cm]{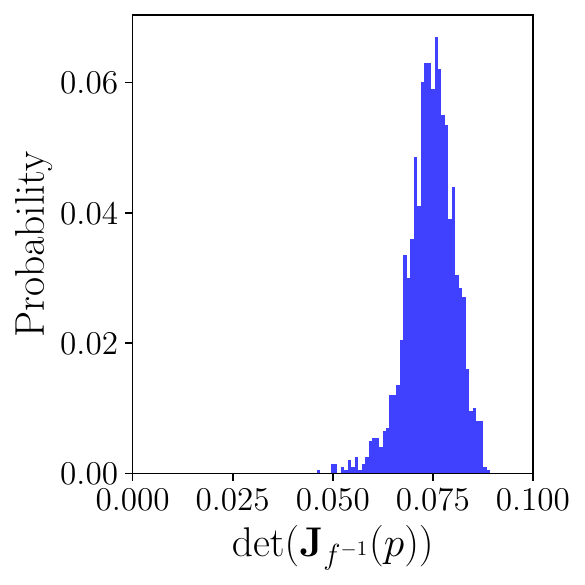}
    \caption{The histogram of the determinant of the Jacobian $\text{det}(\mathbf{J}f^{-1} (\mathbf{p}))$ computed on the test data with symbolic differentiation of the Double DMaps - Geometric Harmonics formula.}
\label{fig:JacobianParameters}
\end{figure}

To ensure that the one-to-one correspondence also holds for the inverse mapping $f^{-1}:\vect{\phi} \mapsto \vect{p}$ we constructed an additional regression scheme based on the Geometric Harmonics algorithm. The number of eigenvectors also in this case was set to 250 and the kernel hyperparameter was computed as the median of the pairwise distances of the rescaled parameters. The absolute errors for a test set of the six non-harmonic coordinates are shown in Figure \ref{GHBoxPhis}.
The  absolute error across the estimated Diffusion Maps coordinates does not exceed $0.03$. In Figure \ref{fig:parityplotsDMAPScoordinates} in the Appendix, we present parity plots comparing the true and predicted values for each individual Diffusion Maps coordinate. The estimated determinant of the Jacobian along the data is also non-singular and consistently positive across the data as shown in Figure \ref{fig:JacobianParameters}.

\section{Discussion}
\label{sec:discussion}

The results presented in this study demonstrate the viability of using data-driven methods for parameter identifiability and reduction.
The synchronous generator model that we have used as a benchmark is a classical model from the power systems domain that has a long-history of analytical studies \cite{sauer1988integral,kokotovic1989integral}.
In particular, the constraints in Eqs.~\eqref{eq:model_constraints} define a hierarchy of time scales that form the basis of singularly perturbed models\cite{kokotovic1976singular}.
In singular perturbation, the model is simplified by taking the fastest time scale, i.e., the smallest $T_i$, to zero, algebraicly constraining the fast degrees of freedom to a slow manifold\cite{petzold1982differential/algebraic}.
Unidentifiable parameters are often those that are coupled to the fastest time scales and similarly removed in the singular limit.
This analysis reflects the reasonable expectation that information about these parameters can only be inferred through a faster initial sampling rate and that these parameters have little impact on the long-term dynamics of interest.

The singular perturbation argument is largely borne out by both the analytic FIM and MBAM analysis as well as the manifold learning analysis.
Indeed, MBAM approximations summarized in Table~\ref{tab:mbam} explicitly include several singularly perturbed limits ($T_{d0}'', T_{q0}'' \rightarrow 0$).
It also explicitly simplifies relationships among fast degrees of freedom by removing the parameter $\delta x_1$.

Surprisingly, however, some of the unidentifiable parameters found by both methods involve the slowest dynamics associated with the ``mechanical'' parameters $D$ (dissipation) and $H$ (rotor moment of inertia).

In traditional singular perturbation analysis, such parameters are considered the most easily identified from data since they are associated with the final transient approach to the steady state.
This apparent contradiction is resolved by inspecting Eqs.~\eqref{eq:model_dyn}.
Observe that parameters $D$ and $H$ both appear in the equation for the rotor speed $\omega$ which is nearly constant in the dynamic response of  Figure~\ref{fig:dynamicresponse}.
The nominal initial condition for $\omega$ selected was chosen already very near its equilibrium value, so its dynamic trajectory is approximately constant and does not carry much information about the parameters $D$ and $H$.
If the initial $\omega$ had been farther from equilibrium, the resulting slow exponential trajectory would have carried more information about the parameters $D$ and $H$, making them more identifiable.

The foregoing observation suggests nuances in the role of initial conditions for transient data collection in nonlinear system identification\cite{saccomani2003parameter}.

%In particular, what type of initial conditions are ``generic'' for models such as this?

In the present study, the initial conditions were constructed by first assuming the system was in steady state.
It was then disrupted by a brief short.
The trajectories in Figure~\ref{fig:dynamicresponse} begin immediately after the short has cleared and depict the relaxation back to the previous steady state.
Because the simulation begins after (or near) a brief disturbance, the slow variables, e.g., $\omega$, have not been perturbed significantly from their steady state and are, consequently, unidentifiable.
In contrast, if our simulation had commenced after, for example, an increase in load that changed the dynamical fixed point, the system would relax to a different fixed point.
In such a case, the parameters associated with the slow variables would have been identifiable.

The preceding discussion illustrates  potential nuances associated with parameter identifiability from real data.
The methods we present here, both data-driven and analytic, can tease out such subtleties and provide useful insights that can guide how system models are identified and applied. 

A striking result of the present analysis is the agreement between both analytic methods (FIM and MBAM) and the data driven methods (Output - Diffusion Maps), which validates the data-driven methods.
Both classes of methods agree on the number of identifiable parameters (six) and which subset of parameters are most identifiable.
This agreement is, in our opinion, non-obvious because each method probes different scales of parameter variations.
In particular, the FIM is based on a local sensitivity analysis, and does not incorporate any nonlinear response to changes in parameters.
On the other hand, MBAM is based on the global boundary structure of the model manifold.
In contrast, the data-driven methods probe intermediate scales (we varied parameters 10\% from their nominal values) where we do not expect closed-form expressions of identifiable parameters to be forthcoming.
In general, one might expect that the identifiable subset of parameters can depend on which scale of variation is considered.
In this regard, we expect data-driven techniques to provide a potentially useful interpolation between the analytical studies of the FIM and the MBAM.
Future studies may use data-driven techniques to study how local identifiability relates to global identifiability.

In addition to probing different scales of parameter variation, the data-driven techniques presented here complement analytic methods in the types of calculations they require.
Critically, analytic methods are based on linearization of the model response.
Sensitivity analysis is central to both FIM and MBAM, as well as other methods such as active subspaces\cite{constantine2015active}.
In contrast, the data driven techniques require only samples of the model dynamics  at several parameter values.
This makes them applicable to a broader class of problems, including models that incorporate multiple code-bases or legacy routines for which automatic differentiation is difficult to apply.

A potential challenge in interpreting our results derives from reparameterizing the model in terms of $\delta x$ and $\delta T$.
Recall that this reparameterization was motivated by the need to satisfy constraints in Eq.~\eqref{eq:model_constraints} while randomly perturbing parameters.
In the present case, it is not hard to work back to express the identifiable combinations in terms of the original parameters.
However, for more complicated constraints, this may be less straightforward.

An open question is how data-driven methods scale to large models.
The curse of dimensionality and the need to thoroughly sample the parameter space could lead to a large computational burden.
Because unidentifiable, sloppy models exhibit a response that is much lower dimension, it may be possible strategically sample to exploit this structure.
%%%YGK the english of the above sentence does not jive - 
Future work could consider hybrid methods that leverage both analytic and data-driven methods to exploit low-dimensional structures for building useful, identifiable models.

The present study has laid the foundation for applying data-driven methods for identifiability analysis of complex, multi-parameter models.
We have used a classic model from power systems as a benchmark example and demonstrated agreement between traditional analytic treatment with emerging techniques based on manifold learning.
We anticipate these methods will be increasingly relevant for high-fidelity multi-scale and multi-physics models underlying emerging technology such as digital twins.

\section{Acknowledgements}

This work was supported by the NSF under award number ECCS-2223987 (NE, IGK) and ECCS-2223986 (AMS), and ECCS-2223985 (MKT).

\bibliographystyle{apsrev4-2}
\bibliography{lit}

\begin{thebibliography}{52}%
\makeatletter
\providecommand \@ifxundefined [1]{%
 \@ifx{#1\undefined}
}%
\providecommand \@ifnum [1]{%
 \ifnum #1\expandafter \@firstoftwo
 \else \expandafter \@secondoftwo
 \fi
}%
\providecommand \@ifx [1]{%
 \ifx #1\expandafter \@firstoftwo
 \else \expandafter \@secondoftwo
 \fi
}%
\providecommand \natexlab [1]{#1}%
\providecommand \enquote  [1]{``#1''}%
\providecommand \bibnamefont  [1]{#1}%
\providecommand \bibfnamefont [1]{#1}%
\providecommand \citenamefont [1]{#1}%
\providecommand \href@noop [0]{\@secondoftwo}%
\providecommand \href [0]{\begingroup \@sanitize@url \@href}%
\providecommand \@href[1]{\@@startlink{#1}\@@href}%
\providecommand \@@href[1]{\endgroup#1\@@endlink}%
\providecommand \@sanitize@url [0]{\catcode `\\12\catcode `\$12\catcode
  `\&12\catcode `\#12\catcode `\^12\catcode `\_12\catcode `\%12\relax}%
\providecommand \@@startlink[1]{}%
\providecommand \@@endlink[0]{}%
\providecommand \url  [0]{\begingroup\@sanitize@url \@url }%
\providecommand \@url [1]{\endgroup\@href {#1}{\urlprefix }}%
\providecommand \urlprefix  [0]{URL }%
\providecommand \Eprint [0]{\href }%
\providecommand \doibase [0]{https://doi.org/}%
\providecommand \selectlanguage [0]{\@gobble}%
\providecommand \bibinfo  [0]{\@secondoftwo}%
\providecommand \bibfield  [0]{\@secondoftwo}%
\providecommand \translation [1]{[#1]}%
\providecommand \BibitemOpen [0]{}%
\providecommand \bibitemStop [0]{}%
\providecommand \bibitemNoStop [0]{.\EOS\space}%
\providecommand \EOS [0]{\spacefactor3000\relax}%
\providecommand \BibitemShut  [1]{\csname bibitem#1\endcsname}%
\let\auto@bib@innerbib\@empty
%</preamble>
\bibitem [{\citenamefont {Juarez}\ \emph {et~al.}(2021)\citenamefont {Juarez},
  \citenamefont {Botti},\ and\ \citenamefont {Giret}}]{juarez2021digital}%
  \BibitemOpen
  \bibfield  {author} {\bibinfo {author} {\bibfnamefont {M.~G.}\ \bibnamefont
  {Juarez}}, \bibinfo {author} {\bibfnamefont {V.~J.}\ \bibnamefont {Botti}},\
  and\ \bibinfo {author} {\bibfnamefont {A.~S.}\ \bibnamefont {Giret}},\ }\href
  {https://doi.org/10.1115/1.4050244} {\bibfield  {journal} {\bibinfo
  {journal} {Journal of Computing and Information Science in Engineering}\
  }\textbf {\bibinfo {volume} {21}},\ \bibinfo {pages} {nil} (\bibinfo {year}
  {2021})}\BibitemShut {NoStop}%
\bibitem [{\citenamefont {Huxoll}\ \emph {et~al.}(2021)\citenamefont {Huxoll},
  \citenamefont {Aldebs}, \citenamefont {Baboli}, \citenamefont {Lehnhoff},\
  and\ \citenamefont {Babazadeh}}]{huxoll2021model}%
  \BibitemOpen
  \bibfield  {author} {\bibinfo {author} {\bibfnamefont {N.}~\bibnamefont
  {Huxoll}}, \bibinfo {author} {\bibfnamefont {M.}~\bibnamefont {Aldebs}},
  \bibinfo {author} {\bibfnamefont {P.~T.}\ \bibnamefont {Baboli}}, \bibinfo
  {author} {\bibfnamefont {S.}~\bibnamefont {Lehnhoff}},\ and\ \bibinfo
  {author} {\bibfnamefont {D.}~\bibnamefont {Babazadeh}},\ }in\ \href
  {https://doi.org/10.1109/sest50973.2021.9543095} {\emph {\bibinfo {booktitle}
  {2021 International Conference on Smart Energy Systems and Technologies
  (SEST)}}}\ (\bibinfo {year} {2021})\ pp.\ \bibinfo {pages} {1--6}\BibitemShut
  {NoStop}%
\bibitem [{\citenamefont {Thelen}\ \emph {et~al.}(2022)\citenamefont {Thelen},
  \citenamefont {Zhang}, \citenamefont {Fink}, \citenamefont {Lu},
  \citenamefont {Ghosh}, \citenamefont {Youn}, \citenamefont {Todd},
  \citenamefont {Mahadevan}, \citenamefont {Hu},\ and\ \citenamefont
  {Hu}}]{thelen2022comprehens}%
  \BibitemOpen
  \bibfield  {author} {\bibinfo {author} {\bibfnamefont {A.}~\bibnamefont
  {Thelen}}, \bibinfo {author} {\bibfnamefont {X.}~\bibnamefont {Zhang}},
  \bibinfo {author} {\bibfnamefont {O.}~\bibnamefont {Fink}}, \bibinfo {author}
  {\bibfnamefont {Y.}~\bibnamefont {Lu}}, \bibinfo {author} {\bibfnamefont
  {S.}~\bibnamefont {Ghosh}}, \bibinfo {author} {\bibfnamefont {B.~D.}\
  \bibnamefont {Youn}}, \bibinfo {author} {\bibfnamefont {M.~D.}\ \bibnamefont
  {Todd}}, \bibinfo {author} {\bibfnamefont {S.}~\bibnamefont {Mahadevan}},
  \bibinfo {author} {\bibfnamefont {C.}~\bibnamefont {Hu}},\ and\ \bibinfo
  {author} {\bibfnamefont {Z.}~\bibnamefont {Hu}},\ }\href
  {https://doi.org/10.1007/s00158-022-03425-4} {\bibfield  {journal} {\bibinfo
  {journal} {Structural and Multidisciplinary Optimization}\ }\textbf {\bibinfo
  {volume} {65}},\ \bibinfo {pages} {354} (\bibinfo {year} {2022})}\BibitemShut
  {NoStop}%
\bibitem [{\citenamefont {Fabiani}\ \emph {et~al.}(2024)\citenamefont
  {Fabiani}, \citenamefont {Evangelou}, \citenamefont {Cui}, \citenamefont
  {Bello-Rivas}, \citenamefont {Martin-Linares}, \citenamefont {Siettos},\ and\
  \citenamefont {Kevrekidis}}]{fabiani2024task}%
  \BibitemOpen
  \bibfield  {author} {\bibinfo {author} {\bibfnamefont {G.}~\bibnamefont
  {Fabiani}}, \bibinfo {author} {\bibfnamefont {N.}~\bibnamefont {Evangelou}},
  \bibinfo {author} {\bibfnamefont {T.}~\bibnamefont {Cui}}, \bibinfo {author}
  {\bibfnamefont {J.~M.}\ \bibnamefont {Bello-Rivas}}, \bibinfo {author}
  {\bibfnamefont {C.~P.}\ \bibnamefont {Martin-Linares}}, \bibinfo {author}
  {\bibfnamefont {C.}~\bibnamefont {Siettos}},\ and\ \bibinfo {author}
  {\bibfnamefont {I.~G.}\ \bibnamefont {Kevrekidis}},\ }\href@noop {}
  {\bibfield  {journal} {\bibinfo  {journal} {Nature communications}\ }\textbf
  {\bibinfo {volume} {15}},\ \bibinfo {pages} {4117} (\bibinfo {year}
  {2024})}\BibitemShut {NoStop}%
\bibitem [{\citenamefont {Karpatne}\ \emph {et~al.}(2017)\citenamefont
  {Karpatne}, \citenamefont {Atluri}, \citenamefont {Faghmous}, \citenamefont
  {Steinbach}, \citenamefont {Banerjee}, \citenamefont {Ganguly}, \citenamefont
  {Shekhar}, \citenamefont {Samatova},\ and\ \citenamefont
  {Kumar}}]{karpatne2017theory}%
  \BibitemOpen
  \bibfield  {author} {\bibinfo {author} {\bibfnamefont {A.}~\bibnamefont
  {Karpatne}}, \bibinfo {author} {\bibfnamefont {G.}~\bibnamefont {Atluri}},
  \bibinfo {author} {\bibfnamefont {J.~H.}\ \bibnamefont {Faghmous}}, \bibinfo
  {author} {\bibfnamefont {M.}~\bibnamefont {Steinbach}}, \bibinfo {author}
  {\bibfnamefont {A.}~\bibnamefont {Banerjee}}, \bibinfo {author}
  {\bibfnamefont {A.}~\bibnamefont {Ganguly}}, \bibinfo {author} {\bibfnamefont
  {S.}~\bibnamefont {Shekhar}}, \bibinfo {author} {\bibfnamefont
  {N.}~\bibnamefont {Samatova}},\ and\ \bibinfo {author} {\bibfnamefont
  {V.}~\bibnamefont {Kumar}},\ }\href@noop {} {\bibfield  {journal} {\bibinfo
  {journal} {IEEE Transactions on Knowledge and Data Engineering}\ }\textbf
  {\bibinfo {volume} {29}},\ \bibinfo {pages} {2318} (\bibinfo {year}
  {2017})}\BibitemShut {NoStop}%
\bibitem [{\citenamefont {Willard}\ \emph {et~al.}(2020)\citenamefont
  {Willard}, \citenamefont {Jia}, \citenamefont {Xu}, \citenamefont
  {Steinbach},\ and\ \citenamefont {Kumar}}]{willard2020integrating}%
  \BibitemOpen
  \bibfield  {author} {\bibinfo {author} {\bibfnamefont {J.}~\bibnamefont
  {Willard}}, \bibinfo {author} {\bibfnamefont {X.}~\bibnamefont {Jia}},
  \bibinfo {author} {\bibfnamefont {S.}~\bibnamefont {Xu}}, \bibinfo {author}
  {\bibfnamefont {M.}~\bibnamefont {Steinbach}},\ and\ \bibinfo {author}
  {\bibfnamefont {V.}~\bibnamefont {Kumar}},\ }\href@noop {} {\bibfield
  {journal} {\bibinfo  {journal} {arXiv preprint arXiv:2003.04919}\ }\textbf
  {\bibinfo {volume} {1}},\ \bibinfo {pages} {1} (\bibinfo {year}
  {2020})}\BibitemShut {NoStop}%
\bibitem [{\citenamefont {Rai}\ and\ \citenamefont
  {Sahu}(2020)}]{rai2020driven}%
  \BibitemOpen
  \bibfield  {author} {\bibinfo {author} {\bibfnamefont {R.}~\bibnamefont
  {Rai}}\ and\ \bibinfo {author} {\bibfnamefont {C.~K.}\ \bibnamefont {Sahu}},\
  }\href@noop {} {\bibfield  {journal} {\bibinfo  {journal} {IEEe Access}\
  }\textbf {\bibinfo {volume} {8}},\ \bibinfo {pages} {71050} (\bibinfo {year}
  {2020})}\BibitemShut {NoStop}%
\bibitem [{\citenamefont {Raissi}\ \emph {et~al.}(2019)\citenamefont {Raissi},
  \citenamefont {Perdikaris},\ and\ \citenamefont
  {Karniadakis}}]{raissi2019physics}%
  \BibitemOpen
  \bibfield  {author} {\bibinfo {author} {\bibfnamefont {M.}~\bibnamefont
  {Raissi}}, \bibinfo {author} {\bibfnamefont {P.}~\bibnamefont {Perdikaris}},\
  and\ \bibinfo {author} {\bibfnamefont {G.}~\bibnamefont {Karniadakis}},\
  }\href {https://doi.org/10.1016/j.jcp.2018.10.045} {\bibfield  {journal}
  {\bibinfo  {journal} {Journal of Computational Physics}\ }\textbf {\bibinfo
  {volume} {378}},\ \bibinfo {pages} {686} (\bibinfo {year}
  {2019})}\BibitemShut {NoStop}%
\bibitem [{\citenamefont {Karniadakis}\ \emph {et~al.}(2021)\citenamefont
  {Karniadakis}, \citenamefont {Kevrekidis}, \citenamefont {Lu}, \citenamefont
  {Perdikaris}, \citenamefont {Wang},\ and\ \citenamefont
  {Yang}}]{karniadakis2021physics}%
  \BibitemOpen
  \bibfield  {author} {\bibinfo {author} {\bibfnamefont {G.~E.}\ \bibnamefont
  {Karniadakis}}, \bibinfo {author} {\bibfnamefont {I.~G.}\ \bibnamefont
  {Kevrekidis}}, \bibinfo {author} {\bibfnamefont {L.}~\bibnamefont {Lu}},
  \bibinfo {author} {\bibfnamefont {P.}~\bibnamefont {Perdikaris}}, \bibinfo
  {author} {\bibfnamefont {S.}~\bibnamefont {Wang}},\ and\ \bibinfo {author}
  {\bibfnamefont {L.}~\bibnamefont {Yang}},\ }\href@noop {} {\bibfield
  {journal} {\bibinfo  {journal} {Nature Reviews Physics}\ }\textbf {\bibinfo
  {volume} {3}},\ \bibinfo {pages} {422} (\bibinfo {year} {2021})}\BibitemShut
  {NoStop}%
\bibitem [{\citenamefont {Holiday}\ \emph {et~al.}(2019)\citenamefont
  {Holiday}, \citenamefont {Kooshkbaghi}, \citenamefont {Bello-Rivas},
  \citenamefont {Gear}, \citenamefont {Zagaris},\ and\ \citenamefont
  {Kevrekidis}}]{holiday2019manifold}%
  \BibitemOpen
  \bibfield  {author} {\bibinfo {author} {\bibfnamefont {A.}~\bibnamefont
  {Holiday}}, \bibinfo {author} {\bibfnamefont {M.}~\bibnamefont
  {Kooshkbaghi}}, \bibinfo {author} {\bibfnamefont {J.~M.}\ \bibnamefont
  {Bello-Rivas}}, \bibinfo {author} {\bibfnamefont {C.~W.}\ \bibnamefont
  {Gear}}, \bibinfo {author} {\bibfnamefont {A.}~\bibnamefont {Zagaris}},\ and\
  \bibinfo {author} {\bibfnamefont {I.~G.}\ \bibnamefont {Kevrekidis}},\
  }\href@noop {} {\bibfield  {journal} {\bibinfo  {journal} {Journal of
  computational physics}\ }\textbf {\bibinfo {volume} {392}},\ \bibinfo {pages}
  {419} (\bibinfo {year} {2019})}\BibitemShut {NoStop}%
\bibitem [{\citenamefont {Evangelou}\ \emph {et~al.}(2022)\citenamefont
  {Evangelou}, \citenamefont {Wichrowski}, \citenamefont {Kevrekidis},
  \citenamefont {Dietrich}, \citenamefont {Kooshkbaghi}, \citenamefont
  {McFann},\ and\ \citenamefont {Kevrekidis}}]{evangelou2022parameter}%
  \BibitemOpen
  \bibfield  {author} {\bibinfo {author} {\bibfnamefont {N.}~\bibnamefont
  {Evangelou}}, \bibinfo {author} {\bibfnamefont {N.~J.}\ \bibnamefont
  {Wichrowski}}, \bibinfo {author} {\bibfnamefont {G.~A.}\ \bibnamefont
  {Kevrekidis}}, \bibinfo {author} {\bibfnamefont {F.}~\bibnamefont
  {Dietrich}}, \bibinfo {author} {\bibfnamefont {M.}~\bibnamefont
  {Kooshkbaghi}}, \bibinfo {author} {\bibfnamefont {S.}~\bibnamefont
  {McFann}},\ and\ \bibinfo {author} {\bibfnamefont {I.~G.}\ \bibnamefont
  {Kevrekidis}},\ }\href@noop {} {\bibfield  {journal} {\bibinfo  {journal}
  {PNAS nexus}\ }\textbf {\bibinfo {volume} {1}},\ \bibinfo {pages} {pgac154}
  (\bibinfo {year} {2022})}\BibitemShut {NoStop}%
\bibitem [{\citenamefont {Cobelli}\ and\ \citenamefont
  {DiStefano}(1980)}]{cobelli1980parameter}%
  \BibitemOpen
  \bibfield  {author} {\bibinfo {author} {\bibfnamefont {C.}~\bibnamefont
  {Cobelli}}\ and\ \bibinfo {author} {\bibfnamefont {J.~J.}\ \bibnamefont
  {DiStefano}},\ }\href {https://doi.org/10.1152/ajpregu.1980.239.1.r7}
  {\bibfield  {journal} {\bibinfo  {journal} {American Journal of
  Physiology-Regulatory, Integrative and Comparative Physiology}\ }\textbf
  {\bibinfo {volume} {239}},\ \bibinfo {pages} {R7} (\bibinfo {year}
  {1980})}\BibitemShut {NoStop}%
\bibitem [{\citenamefont {Raue}\ \emph {et~al.}(2014)\citenamefont {Raue},
  \citenamefont {Karlsson}, \citenamefont {Saccomani}, \citenamefont
  {Jirstrand},\ and\ \citenamefont {Timmer}}]{raue2014comparison}%
  \BibitemOpen
  \bibfield  {author} {\bibinfo {author} {\bibfnamefont {A.}~\bibnamefont
  {Raue}}, \bibinfo {author} {\bibfnamefont {J.}~\bibnamefont {Karlsson}},
  \bibinfo {author} {\bibfnamefont {M.~P.}\ \bibnamefont {Saccomani}}, \bibinfo
  {author} {\bibfnamefont {M.}~\bibnamefont {Jirstrand}},\ and\ \bibinfo
  {author} {\bibfnamefont {J.}~\bibnamefont {Timmer}},\ }\href
  {https://doi.org/10.1093/bioinformatics/btu006} {\bibfield  {journal}
  {\bibinfo  {journal} {Bioinformatics}\ }\textbf {\bibinfo {volume} {30}},\
  \bibinfo {pages} {1440} (\bibinfo {year} {2014})}\BibitemShut {NoStop}%
\bibitem [{\citenamefont {Zhan}\ and\ \citenamefont
  {Alcantud}(2017)}]{zhan2017survey}%
  \BibitemOpen
  \bibfield  {author} {\bibinfo {author} {\bibfnamefont {J.}~\bibnamefont
  {Zhan}}\ and\ \bibinfo {author} {\bibfnamefont {J.~C.~R.}\ \bibnamefont
  {Alcantud}},\ }\href {https://doi.org/10.1007/s10462-017-9592-0} {\bibfield
  {journal} {\bibinfo  {journal} {Artificial Intelligence Review}\ }\textbf
  {\bibinfo {volume} {52}},\ \bibinfo {pages} {1839} (\bibinfo {year}
  {2017})}\BibitemShut {NoStop}%
\bibitem [{\citenamefont {Rothenberg}(1971)}]{rothenberg1971identificat}%
  \BibitemOpen
  \bibfield  {author} {\bibinfo {author} {\bibfnamefont {T.~J.}\ \bibnamefont
  {Rothenberg}},\ }\href {https://doi.org/10.2307/1913267} {\bibfield
  {journal} {\bibinfo  {journal} {Econometrica}\ }\textbf {\bibinfo {volume}
  {39}},\ \bibinfo {pages} {577} (\bibinfo {year} {1971})}\BibitemShut
  {NoStop}%
\bibitem [{\citenamefont {Brouwer}\ and\ \citenamefont
  {Eisenberg}(2018)}]{brouwer2018underlying}%
  \BibitemOpen
  \bibfield  {author} {\bibinfo {author} {\bibfnamefont {A.~F.}\ \bibnamefont
  {Brouwer}}\ and\ \bibinfo {author} {\bibfnamefont {M.~C.}\ \bibnamefont
  {Eisenberg}},\ }\href@noop {} {\bibfield  {journal} {\bibinfo  {journal}
  {arXiv preprint arXiv:1802.05641}\ } (\bibinfo {year} {2018})}\BibitemShut
  {NoStop}%
\bibitem [{\citenamefont {Brown}\ \emph {et~al.}(2004)\citenamefont {Brown},
  \citenamefont {Hill}, \citenamefont {Calero}, \citenamefont {Myers},
  \citenamefont {Lee}, \citenamefont {Sethna},\ and\ \citenamefont
  {Cerione}}]{brown2004statistical}%
  \BibitemOpen
  \bibfield  {author} {\bibinfo {author} {\bibfnamefont {K.~S.}\ \bibnamefont
  {Brown}}, \bibinfo {author} {\bibfnamefont {C.~C.}\ \bibnamefont {Hill}},
  \bibinfo {author} {\bibfnamefont {G.~A.}\ \bibnamefont {Calero}}, \bibinfo
  {author} {\bibfnamefont {C.~R.}\ \bibnamefont {Myers}}, \bibinfo {author}
  {\bibfnamefont {K.~H.}\ \bibnamefont {Lee}}, \bibinfo {author} {\bibfnamefont
  {J.~P.}\ \bibnamefont {Sethna}},\ and\ \bibinfo {author} {\bibfnamefont
  {R.~A.}\ \bibnamefont {Cerione}},\ }\href
  {https://doi.org/10.1088/1478-3967/1/3/006} {\bibfield  {journal} {\bibinfo
  {journal} {Physical Biology}\ }\textbf {\bibinfo {volume} {1}},\ \bibinfo
  {pages} {184} (\bibinfo {year} {2004})}\BibitemShut {NoStop}%
\bibitem [{\citenamefont {Gutenkunst}\ \emph {et~al.}(2007)\citenamefont
  {Gutenkunst}, \citenamefont {Waterfall}, \citenamefont {Casey}, \citenamefont
  {Brown}, \citenamefont {Myers},\ and\ \citenamefont
  {Sethna}}]{gutenkunst2007universally}%
  \BibitemOpen
  \bibfield  {author} {\bibinfo {author} {\bibfnamefont {R.~N.}\ \bibnamefont
  {Gutenkunst}}, \bibinfo {author} {\bibfnamefont {J.~J.}\ \bibnamefont
  {Waterfall}}, \bibinfo {author} {\bibfnamefont {F.~P.}\ \bibnamefont
  {Casey}}, \bibinfo {author} {\bibfnamefont {K.~S.}\ \bibnamefont {Brown}},
  \bibinfo {author} {\bibfnamefont {C.~R.}\ \bibnamefont {Myers}},\ and\
  \bibinfo {author} {\bibfnamefont {J.~P.}\ \bibnamefont {Sethna}},\ }\href
  {https://doi.org/10.1371/journal.pcbi.0030189} {\bibfield  {journal}
  {\bibinfo  {journal} {PLoS Computational Biology}\ }\textbf {\bibinfo
  {volume} {3}},\ \bibinfo {pages} {e189} (\bibinfo {year} {2007})}\BibitemShut
  {NoStop}%
\bibitem [{\citenamefont {Transtrum}\ \emph {et~al.}(2015)\citenamefont
  {Transtrum}, \citenamefont {Machta}, \citenamefont {Brown}, \citenamefont
  {Daniels}, \citenamefont {Myers},\ and\ \citenamefont
  {Sethna}}]{transtrum2015perspective}%
  \BibitemOpen
  \bibfield  {author} {\bibinfo {author} {\bibfnamefont {M.~K.}\ \bibnamefont
  {Transtrum}}, \bibinfo {author} {\bibfnamefont {B.~B.}\ \bibnamefont
  {Machta}}, \bibinfo {author} {\bibfnamefont {K.~S.}\ \bibnamefont {Brown}},
  \bibinfo {author} {\bibfnamefont {B.~C.}\ \bibnamefont {Daniels}}, \bibinfo
  {author} {\bibfnamefont {C.~R.}\ \bibnamefont {Myers}},\ and\ \bibinfo
  {author} {\bibfnamefont {J.~P.}\ \bibnamefont {Sethna}},\ }\href
  {https://doi.org/10.1063/1.4923066} {\bibfield  {journal} {\bibinfo
  {journal} {The Journal of Chemical Physics}\ }\textbf {\bibinfo {volume}
  {143}},\ \bibinfo {pages} {010901} (\bibinfo {year} {2015})}\BibitemShut
  {NoStop}%
\bibitem [{\citenamefont {Quinn}\ \emph {et~al.}(2022)\citenamefont {Quinn},
  \citenamefont {Abbott}, \citenamefont {Transtrum}, \citenamefont {Machta},\
  and\ \citenamefont {Sethna}}]{quinn2022information}%
  \BibitemOpen
  \bibfield  {author} {\bibinfo {author} {\bibfnamefont {K.~N.}\ \bibnamefont
  {Quinn}}, \bibinfo {author} {\bibfnamefont {M.~C.}\ \bibnamefont {Abbott}},
  \bibinfo {author} {\bibfnamefont {M.~K.}\ \bibnamefont {Transtrum}}, \bibinfo
  {author} {\bibfnamefont {B.~B.}\ \bibnamefont {Machta}},\ and\ \bibinfo
  {author} {\bibfnamefont {J.~P.}\ \bibnamefont {Sethna}},\ }\href@noop {}
  {\bibfield  {journal} {\bibinfo  {journal} {Reports on Progress in Physics}\
  }\textbf {\bibinfo {volume} {86}},\ \bibinfo {pages} {035901} (\bibinfo
  {year} {2022})}\BibitemShut {NoStop}%
\bibitem [{\citenamefont {Frederiksen}\ \emph {et~al.}(2004)\citenamefont
  {Frederiksen}, \citenamefont {Jacobsen}, \citenamefont {Brown},\ and\
  \citenamefont {Sethna}}]{frederiksen2004bayesian}%
  \BibitemOpen
  \bibfield  {author} {\bibinfo {author} {\bibfnamefont {S.~L.}\ \bibnamefont
  {Frederiksen}}, \bibinfo {author} {\bibfnamefont {K.~W.}\ \bibnamefont
  {Jacobsen}}, \bibinfo {author} {\bibfnamefont {K.~S.}\ \bibnamefont
  {Brown}},\ and\ \bibinfo {author} {\bibfnamefont {J.~P.}\ \bibnamefont
  {Sethna}},\ }\href {https://doi.org/10.1103/physrevlett.93.165501} {\bibfield
   {journal} {\bibinfo  {journal} {Physical Review Letters}\ }\textbf {\bibinfo
  {volume} {93}},\ \bibinfo {pages} {165501} (\bibinfo {year}
  {2004})}\BibitemShut {NoStop}%
\bibitem [{\citenamefont {Waterfall}\ \emph {et~al.}(2006)\citenamefont
  {Waterfall}, \citenamefont {Casey}, \citenamefont {Gutenkunst}, \citenamefont
  {Brown}, \citenamefont {Myers}, \citenamefont {Brouwer}, \citenamefont
  {Elser},\ and\ \citenamefont {Sethna}}]{waterfall2006sloppy}%
  \BibitemOpen
  \bibfield  {author} {\bibinfo {author} {\bibfnamefont {J.~J.}\ \bibnamefont
  {Waterfall}}, \bibinfo {author} {\bibfnamefont {F.~P.}\ \bibnamefont
  {Casey}}, \bibinfo {author} {\bibfnamefont {R.~N.}\ \bibnamefont
  {Gutenkunst}}, \bibinfo {author} {\bibfnamefont {K.~S.}\ \bibnamefont
  {Brown}}, \bibinfo {author} {\bibfnamefont {C.~R.}\ \bibnamefont {Myers}},
  \bibinfo {author} {\bibfnamefont {P.~W.}\ \bibnamefont {Brouwer}}, \bibinfo
  {author} {\bibfnamefont {V.}~\bibnamefont {Elser}},\ and\ \bibinfo {author}
  {\bibfnamefont {J.~P.}\ \bibnamefont {Sethna}},\ }\href
  {https://doi.org/10.1103/physrevlett.97.150601} {\bibfield  {journal}
  {\bibinfo  {journal} {Physical Review Letters}\ }\textbf {\bibinfo {volume}
  {97}},\ \bibinfo {pages} {150601} (\bibinfo {year} {2006})}\BibitemShut
  {NoStop}%
\bibitem [{\citenamefont {Transtrum}\ \emph {et~al.}(2018)\citenamefont
  {Transtrum}, \citenamefont {Saric},\ and\ \citenamefont
  {Stankovic}}]{transtrum2018information}%
  \BibitemOpen
  \bibfield  {author} {\bibinfo {author} {\bibfnamefont {M.~K.}\ \bibnamefont
  {Transtrum}}, \bibinfo {author} {\bibfnamefont {A.~T.}\ \bibnamefont
  {Saric}},\ and\ \bibinfo {author} {\bibfnamefont {A.~M.}\ \bibnamefont
  {Stankovic}},\ }\href {https://doi.org/10.1109/tpwrs.2017.2692523} {\bibfield
   {journal} {\bibinfo  {journal} {IEEE Transactions on Power Systems}\
  }\textbf {\bibinfo {volume} {33}},\ \bibinfo {pages} {440} (\bibinfo {year}
  {2018})}\BibitemShut {NoStop}%
\bibitem [{\citenamefont {Machta}\ \emph {et~al.}(2013)\citenamefont {Machta},
  \citenamefont {Chachra}, \citenamefont {Transtrum},\ and\ \citenamefont
  {Sethna}}]{machta2013parameter}%
  \BibitemOpen
  \bibfield  {author} {\bibinfo {author} {\bibfnamefont {B.~B.}\ \bibnamefont
  {Machta}}, \bibinfo {author} {\bibfnamefont {R.}~\bibnamefont {Chachra}},
  \bibinfo {author} {\bibfnamefont {M.~K.}\ \bibnamefont {Transtrum}},\ and\
  \bibinfo {author} {\bibfnamefont {J.~P.}\ \bibnamefont {Sethna}},\ }\href
  {https://doi.org/10.1126/science.1238723} {\bibfield  {journal} {\bibinfo
  {journal} {Science}\ }\textbf {\bibinfo {volume} {342}},\ \bibinfo {pages}
  {604} (\bibinfo {year} {2013})}\BibitemShut {NoStop}%
\bibitem [{\citenamefont {Transtrum}\ \emph {et~al.}(2010)\citenamefont
  {Transtrum}, \citenamefont {Machta},\ and\ \citenamefont
  {Sethna}}]{transtrum2010why}%
  \BibitemOpen
  \bibfield  {author} {\bibinfo {author} {\bibfnamefont {M.~K.}\ \bibnamefont
  {Transtrum}}, \bibinfo {author} {\bibfnamefont {B.~B.}\ \bibnamefont
  {Machta}},\ and\ \bibinfo {author} {\bibfnamefont {J.~P.}\ \bibnamefont
  {Sethna}},\ }\href {https://doi.org/10.1103/physrevlett.104.060201}
  {\bibfield  {journal} {\bibinfo  {journal} {Physical Review Letters}\
  }\textbf {\bibinfo {volume} {104}},\ \bibinfo {pages} {060201} (\bibinfo
  {year} {2010})}\BibitemShut {NoStop}%
\bibitem [{\citenamefont {Transtrum}\ \emph {et~al.}(2011)\citenamefont
  {Transtrum}, \citenamefont {Machta},\ and\ \citenamefont
  {Sethna}}]{transtrum2011geometry}%
  \BibitemOpen
  \bibfield  {author} {\bibinfo {author} {\bibfnamefont {M.~K.}\ \bibnamefont
  {Transtrum}}, \bibinfo {author} {\bibfnamefont {B.~B.}\ \bibnamefont
  {Machta}},\ and\ \bibinfo {author} {\bibfnamefont {J.~P.}\ \bibnamefont
  {Sethna}},\ }\href {https://doi.org/10.1103/physreve.83.036701} {\bibfield
  {journal} {\bibinfo  {journal} {Physical Review E}\ }\textbf {\bibinfo
  {volume} {83}},\ \bibinfo {pages} {036701} (\bibinfo {year}
  {2011})}\BibitemShut {NoStop}%
\bibitem [{\citenamefont {Quinn}\ \emph
  {et~al.}(2019{\natexlab{a}})\citenamefont {Quinn}, \citenamefont {Clement},
  \citenamefont {De~Bernardis}, \citenamefont {Niemack},\ and\ \citenamefont
  {Sethna}}]{quinn2019visualizing}%
  \BibitemOpen
  \bibfield  {author} {\bibinfo {author} {\bibfnamefont {K.~N.}\ \bibnamefont
  {Quinn}}, \bibinfo {author} {\bibfnamefont {C.~B.}\ \bibnamefont {Clement}},
  \bibinfo {author} {\bibfnamefont {F.}~\bibnamefont {De~Bernardis}}, \bibinfo
  {author} {\bibfnamefont {M.~D.}\ \bibnamefont {Niemack}},\ and\ \bibinfo
  {author} {\bibfnamefont {J.~P.}\ \bibnamefont {Sethna}},\ }\href@noop {}
  {\bibfield  {journal} {\bibinfo  {journal} {Proceedings of the National
  Academy of Sciences}\ }\textbf {\bibinfo {volume} {116}},\ \bibinfo {pages}
  {13762} (\bibinfo {year} {2019}{\natexlab{a}})}\BibitemShut {NoStop}%
\bibitem [{\citenamefont {Quinn}\ \emph
  {et~al.}(2019{\natexlab{b}})\citenamefont {Quinn}, \citenamefont {Wilber},
  \citenamefont {Townsend},\ and\ \citenamefont {Sethna}}]{quinn2019chebyshev}%
  \BibitemOpen
  \bibfield  {author} {\bibinfo {author} {\bibfnamefont {K.~N.}\ \bibnamefont
  {Quinn}}, \bibinfo {author} {\bibfnamefont {H.}~\bibnamefont {Wilber}},
  \bibinfo {author} {\bibfnamefont {A.}~\bibnamefont {Townsend}},\ and\
  \bibinfo {author} {\bibfnamefont {J.~P.}\ \bibnamefont {Sethna}},\
  }\href@noop {} {\bibfield  {journal} {\bibinfo  {journal} {Physical Review
  Letters}\ }\textbf {\bibinfo {volume} {122}},\ \bibinfo {pages} {158302}
  (\bibinfo {year} {2019}{\natexlab{b}})}\BibitemShut {NoStop}%
\bibitem [{\citenamefont {Mattingly}\ \emph {et~al.}(2018)\citenamefont
  {Mattingly}, \citenamefont {Transtrum}, \citenamefont {Abbott},\ and\
  \citenamefont {Machta}}]{mattingly2018maximizing}%
  \BibitemOpen
  \bibfield  {author} {\bibinfo {author} {\bibfnamefont {H.~H.}\ \bibnamefont
  {Mattingly}}, \bibinfo {author} {\bibfnamefont {M.~K.}\ \bibnamefont
  {Transtrum}}, \bibinfo {author} {\bibfnamefont {M.~C.}\ \bibnamefont
  {Abbott}},\ and\ \bibinfo {author} {\bibfnamefont {B.~B.}\ \bibnamefont
  {Machta}},\ }\href@noop {} {\bibfield  {journal} {\bibinfo  {journal}
  {Proceedings of the National Academy of Sciences}\ }\textbf {\bibinfo
  {volume} {115}},\ \bibinfo {pages} {1760} (\bibinfo {year}
  {2018})}\BibitemShut {NoStop}%
\bibitem [{\citenamefont {Box}(1976)}]{box1976science}%
  \BibitemOpen
  \bibfield  {author} {\bibinfo {author} {\bibfnamefont {G.~E.~P.}\
  \bibnamefont {Box}},\ }\href {https://doi.org/10.1080/01621459.1976.10480949}
  {\bibfield  {journal} {\bibinfo  {journal} {Journal of the American
  Statistical Association}\ }\textbf {\bibinfo {volume} {71}},\ \bibinfo
  {pages} {791} (\bibinfo {year} {1976})}\BibitemShut {NoStop}%
\bibitem [{\citenamefont {Goldenfeld}(1999)}]{goldenfeld1999simple}%
  \BibitemOpen
  \bibfield  {author} {\bibinfo {author} {\bibfnamefont {N.}~\bibnamefont
  {Goldenfeld}},\ }\href {https://doi.org/10.1126/science.284.5411.87}
  {\bibfield  {journal} {\bibinfo  {journal} {Science}\ }\textbf {\bibinfo
  {volume} {284}},\ \bibinfo {pages} {87} (\bibinfo {year} {1999})}\BibitemShut
  {NoStop}%
\bibitem [{\citenamefont {Gunning}\ \emph {et~al.}(2019)\citenamefont
  {Gunning}, \citenamefont {Stefik}, \citenamefont {Choi}, \citenamefont
  {Miller}, \citenamefont {Stumpf},\ and\ \citenamefont
  {Yang}}]{gunning2019xai}%
  \BibitemOpen
  \bibfield  {author} {\bibinfo {author} {\bibfnamefont {D.}~\bibnamefont
  {Gunning}}, \bibinfo {author} {\bibfnamefont {M.}~\bibnamefont {Stefik}},
  \bibinfo {author} {\bibfnamefont {J.}~\bibnamefont {Choi}}, \bibinfo {author}
  {\bibfnamefont {T.}~\bibnamefont {Miller}}, \bibinfo {author} {\bibfnamefont
  {S.}~\bibnamefont {Stumpf}},\ and\ \bibinfo {author} {\bibfnamefont {G.-Z.}\
  \bibnamefont {Yang}},\ }\href {https://doi.org/10.1126/scirobotics.aay7120}
  {\bibfield  {journal} {\bibinfo  {journal} {Science Robotics}\ }\textbf
  {\bibinfo {volume} {4}},\ \bibinfo {pages} {nil} (\bibinfo {year}
  {2019})}\BibitemShut {NoStop}%
\bibitem [{\citenamefont {Transtrum}\ and\ \citenamefont
  {Qiu}(2014)}]{transtrum2014model}%
  \BibitemOpen
  \bibfield  {author} {\bibinfo {author} {\bibfnamefont {M.~K.}\ \bibnamefont
  {Transtrum}}\ and\ \bibinfo {author} {\bibfnamefont {P.}~\bibnamefont
  {Qiu}},\ }\href {https://doi.org/10.1103/physrevlett.113.098701} {\bibfield
  {journal} {\bibinfo  {journal} {Physical Review Letters}\ }\textbf {\bibinfo
  {volume} {113}},\ \bibinfo {pages} {098701} (\bibinfo {year}
  {2014})}\BibitemShut {NoStop}%
\bibitem [{\citenamefont {Transtrum}\ and\ \citenamefont
  {Qiu}(2016)}]{transtrum2016bridging}%
  \BibitemOpen
  \bibfield  {author} {\bibinfo {author} {\bibfnamefont {M.~K.}\ \bibnamefont
  {Transtrum}}\ and\ \bibinfo {author} {\bibfnamefont {P.}~\bibnamefont
  {Qiu}},\ }\href {https://doi.org/10.1371/journal.pcbi.1004915} {\bibfield
  {journal} {\bibinfo  {journal} {PLOS Computational Biology}\ }\textbf
  {\bibinfo {volume} {12}},\ \bibinfo {pages} {e1004915} (\bibinfo {year}
  {2016})}\BibitemShut {NoStop}%
\bibitem [{\citenamefont {Transtrum}\ \emph {et~al.}(2017)\citenamefont
  {Transtrum}, \citenamefont {Saric},\ and\ \citenamefont
  {Stankovic}}]{transtrum2017measurement}%
  \BibitemOpen
  \bibfield  {author} {\bibinfo {author} {\bibfnamefont {M.~K.}\ \bibnamefont
  {Transtrum}}, \bibinfo {author} {\bibfnamefont {A.~T.}\ \bibnamefont
  {Saric}},\ and\ \bibinfo {author} {\bibfnamefont {A.~M.}\ \bibnamefont
  {Stankovic}},\ }\href {https://doi.org/10.1109/tpwrs.2016.2611511} {\bibfield
   {journal} {\bibinfo  {journal} {IEEE Transactions on Power Systems}\
  }\textbf {\bibinfo {volume} {32}},\ \bibinfo {pages} {2243} (\bibinfo {year}
  {2017})}\BibitemShut {NoStop}%
\bibitem [{\citenamefont {Constantine}(2015)}]{constantine2015active}%
  \BibitemOpen
  \bibfield  {author} {\bibinfo {author} {\bibfnamefont {P.~G.}\ \bibnamefont
  {Constantine}},\ }\href@noop {} {\emph {\bibinfo {title} {Active subspaces:
  Emerging ideas for dimension reduction in parameter studies}}}\ (\bibinfo
  {publisher} {SIAM},\ \bibinfo {year} {2015})\BibitemShut {NoStop}%
\bibitem [{\citenamefont {Peterfreund}\ \emph {et~al.}(2020)\citenamefont
  {Peterfreund}, \citenamefont {Lindenbaum}, \citenamefont {Dietrich},
  \citenamefont {Bertalan}, \citenamefont {Gavish}, \citenamefont
  {Kevrekidis},\ and\ \citenamefont {Coifman}}]{peterfreund2020local}%
  \BibitemOpen
  \bibfield  {author} {\bibinfo {author} {\bibfnamefont {E.}~\bibnamefont
  {Peterfreund}}, \bibinfo {author} {\bibfnamefont {O.}~\bibnamefont
  {Lindenbaum}}, \bibinfo {author} {\bibfnamefont {F.}~\bibnamefont
  {Dietrich}}, \bibinfo {author} {\bibfnamefont {T.}~\bibnamefont {Bertalan}},
  \bibinfo {author} {\bibfnamefont {M.}~\bibnamefont {Gavish}}, \bibinfo
  {author} {\bibfnamefont {I.~G.}\ \bibnamefont {Kevrekidis}},\ and\ \bibinfo
  {author} {\bibfnamefont {R.~R.}\ \bibnamefont {Coifman}},\ }\href@noop {}
  {\bibfield  {journal} {\bibinfo  {journal} {Proceedings of the National
  Academy of Sciences}\ }\textbf {\bibinfo {volume} {117}},\ \bibinfo {pages}
  {30918} (\bibinfo {year} {2020})}\BibitemShut {NoStop}%
\bibitem [{\citenamefont {Andersson}\ \emph {et~al.}(2005)\citenamefont
  {Andersson}, \citenamefont {Donalek}, \citenamefont {Farmer}, \citenamefont
  {Hatziargyriou}, \citenamefont {Kamwa}, \citenamefont {Kundur}, \citenamefont
  {Martins}, \citenamefont {Paserba}, \citenamefont {Pourbeik}, \citenamefont
  {Sanchez-Gasca} \emph {et~al.}}]{andersson2005causes}%
  \BibitemOpen
  \bibfield  {author} {\bibinfo {author} {\bibfnamefont {G.}~\bibnamefont
  {Andersson}}, \bibinfo {author} {\bibfnamefont {P.}~\bibnamefont {Donalek}},
  \bibinfo {author} {\bibfnamefont {R.}~\bibnamefont {Farmer}}, \bibinfo
  {author} {\bibfnamefont {N.}~\bibnamefont {Hatziargyriou}}, \bibinfo {author}
  {\bibfnamefont {I.}~\bibnamefont {Kamwa}}, \bibinfo {author} {\bibfnamefont
  {P.}~\bibnamefont {Kundur}}, \bibinfo {author} {\bibfnamefont
  {N.}~\bibnamefont {Martins}}, \bibinfo {author} {\bibfnamefont
  {J.}~\bibnamefont {Paserba}}, \bibinfo {author} {\bibfnamefont
  {P.}~\bibnamefont {Pourbeik}}, \bibinfo {author} {\bibfnamefont
  {J.}~\bibnamefont {Sanchez-Gasca}}, \emph {et~al.},\ }\href@noop {}
  {\bibfield  {journal} {\bibinfo  {journal} {IEEE transactions on Power
  Systems}\ }\textbf {\bibinfo {volume} {20}},\ \bibinfo {pages} {1922}
  (\bibinfo {year} {2005})}\BibitemShut {NoStop}%
\bibitem [{\citenamefont {Sauer}\ \emph {et~al.}(2017)\citenamefont {Sauer},
  \citenamefont {Pai},\ and\ \citenamefont {Chow}}]{sauer2017power}%
  \BibitemOpen
  \bibfield  {author} {\bibinfo {author} {\bibfnamefont {P.~W.}\ \bibnamefont
  {Sauer}}, \bibinfo {author} {\bibfnamefont {M.~A.}\ \bibnamefont {Pai}},\
  and\ \bibinfo {author} {\bibfnamefont {J.~H.}\ \bibnamefont {Chow}},\ }\href
  {https://onlinelibrary.wiley.com/doi/book/10.1002/9781119355755} {\emph
  {\bibinfo {title} {Power system dynamics and stability: with synchrophasor
  measurement and power system toolbox}}}\ (\bibinfo  {publisher} {John Wiley
  \& Sons},\ \bibinfo {year} {2017})\BibitemShut {NoStop}%
\bibitem [{\citenamefont {Revels}\ \emph {et~al.}(2016)\citenamefont {Revels},
  \citenamefont {Lubin},\ and\ \citenamefont {Papamarkou}}]{revels2016forward}%
  \BibitemOpen
  \bibfield  {author} {\bibinfo {author} {\bibfnamefont {J.}~\bibnamefont
  {Revels}}, \bibinfo {author} {\bibfnamefont {M.}~\bibnamefont {Lubin}},\ and\
  \bibinfo {author} {\bibfnamefont {T.}~\bibnamefont {Papamarkou}},\ }\href
  {http://arxiv.org/abs/1607.07892v1} {\bibfield  {journal} {\bibinfo
  {journal} {CoRR}\ } (\bibinfo {year} {2016})},\ \Eprint
  {https://arxiv.org/abs/1607.07892} {arXiv:1607.07892 [cs.MS]} \BibitemShut
  {NoStop}%
\bibitem [{\citenamefont {Transtrum}\ \emph {et~al.}(2014)\citenamefont
  {Transtrum}, \citenamefont {Hart},\ and\ \citenamefont
  {Qiu}}]{transtrum2014information}%
  \BibitemOpen
  \bibfield  {author} {\bibinfo {author} {\bibfnamefont {M.~K.}\ \bibnamefont
  {Transtrum}}, \bibinfo {author} {\bibfnamefont {G.}~\bibnamefont {Hart}},\
  and\ \bibinfo {author} {\bibfnamefont {P.}~\bibnamefont {Qiu}},\ }\href
  {http://arxiv.org/abs/1409.6203v2} {\bibfield  {journal} {\bibinfo  {journal}
  {CoRR}\ } (\bibinfo {year} {2014})},\ \Eprint
  {https://arxiv.org/abs/1409.6203} {arXiv:1409.6203 [physics.data-an]}
  \BibitemShut {NoStop}%
\bibitem [{\citenamefont {Dsilva}\ \emph {et~al.}(2018)\citenamefont {Dsilva},
  \citenamefont {Talmon}, \citenamefont {Coifman},\ and\ \citenamefont
  {Kevrekidis}}]{dsilva2018parsimonious}%
  \BibitemOpen
  \bibfield  {author} {\bibinfo {author} {\bibfnamefont {C.~J.}\ \bibnamefont
  {Dsilva}}, \bibinfo {author} {\bibfnamefont {R.}~\bibnamefont {Talmon}},
  \bibinfo {author} {\bibfnamefont {R.~R.}\ \bibnamefont {Coifman}},\ and\
  \bibinfo {author} {\bibfnamefont {I.~G.}\ \bibnamefont {Kevrekidis}},\
  }\href@noop {} {\bibfield  {journal} {\bibinfo  {journal} {Applied and
  Computational Harmonic Analysis}\ }\textbf {\bibinfo {volume} {44}},\
  \bibinfo {pages} {759} (\bibinfo {year} {2018})}\BibitemShut {NoStop}%
\bibitem [{\citenamefont {Coifman}\ and\ \citenamefont
  {Lafon}(2006)}]{coifman2006geometric}%
  \BibitemOpen
  \bibfield  {author} {\bibinfo {author} {\bibfnamefont {R.~R.}\ \bibnamefont
  {Coifman}}\ and\ \bibinfo {author} {\bibfnamefont {S.}~\bibnamefont
  {Lafon}},\ }\href@noop {} {\bibfield  {journal} {\bibinfo  {journal} {Applied
  and Computational Harmonic Analysis}\ }\textbf {\bibinfo {volume} {21}},\
  \bibinfo {pages} {31} (\bibinfo {year} {2006})}\BibitemShut {NoStop}%
\bibitem [{\citenamefont {Evangelou}\ \emph {et~al.}(2023)\citenamefont
  {Evangelou}, \citenamefont {Dietrich}, \citenamefont {Chiavazzo},
  \citenamefont {Lehmberg}, \citenamefont {Meila},\ and\ \citenamefont
  {Kevrekidis}}]{evangelou2023double}%
  \BibitemOpen
  \bibfield  {author} {\bibinfo {author} {\bibfnamefont {N.}~\bibnamefont
  {Evangelou}}, \bibinfo {author} {\bibfnamefont {F.}~\bibnamefont {Dietrich}},
  \bibinfo {author} {\bibfnamefont {E.}~\bibnamefont {Chiavazzo}}, \bibinfo
  {author} {\bibfnamefont {D.}~\bibnamefont {Lehmberg}}, \bibinfo {author}
  {\bibfnamefont {M.}~\bibnamefont {Meila}},\ and\ \bibinfo {author}
  {\bibfnamefont {I.~G.}\ \bibnamefont {Kevrekidis}},\ }\href@noop {}
  {\bibfield  {journal} {\bibinfo  {journal} {Journal of Computational
  Physics}\ }\textbf {\bibinfo {volume} {485}},\ \bibinfo {pages} {112072}
  (\bibinfo {year} {2023})}\BibitemShut {NoStop}%
\bibitem [{\citenamefont {Koronaki}\ \emph {et~al.}(2023)\citenamefont
  {Koronaki}, \citenamefont {Evangelou}, \citenamefont {Psarellis},
  \citenamefont {Boudouvis},\ and\ \citenamefont
  {Kevrekidis}}]{koronaki2023partial}%
  \BibitemOpen
  \bibfield  {author} {\bibinfo {author} {\bibfnamefont {E.~D.}\ \bibnamefont
  {Koronaki}}, \bibinfo {author} {\bibfnamefont {N.}~\bibnamefont {Evangelou}},
  \bibinfo {author} {\bibfnamefont {Y.~M.}\ \bibnamefont {Psarellis}}, \bibinfo
  {author} {\bibfnamefont {A.~G.}\ \bibnamefont {Boudouvis}},\ and\ \bibinfo
  {author} {\bibfnamefont {I.~G.}\ \bibnamefont {Kevrekidis}},\ }\href@noop {}
  {\bibfield  {journal} {\bibinfo  {journal} {Computers \& Chemical
  Engineering}\ }\textbf {\bibinfo {volume} {178}},\ \bibinfo {pages} {108357}
  (\bibinfo {year} {2023})}\BibitemShut {NoStop}%
\bibitem [{\citenamefont {Koronaki}\ \emph {et~al.}(2024)\citenamefont
  {Koronaki}, \citenamefont {Evangelou}, \citenamefont {Martin-Linares},
  \citenamefont {Titi},\ and\ \citenamefont
  {Kevrekidis}}]{koronaki2024nonlinear}%
  \BibitemOpen
  \bibfield  {author} {\bibinfo {author} {\bibfnamefont {E.~D.}\ \bibnamefont
  {Koronaki}}, \bibinfo {author} {\bibfnamefont {N.}~\bibnamefont {Evangelou}},
  \bibinfo {author} {\bibfnamefont {C.~P.}\ \bibnamefont {Martin-Linares}},
  \bibinfo {author} {\bibfnamefont {E.~S.}\ \bibnamefont {Titi}},\ and\
  \bibinfo {author} {\bibfnamefont {I.~G.}\ \bibnamefont {Kevrekidis}},\
  }\href@noop {} {\bibfield  {journal} {\bibinfo  {journal} {Journal of
  Computational Physics}\ }\textbf {\bibinfo {volume} {506}},\ \bibinfo {pages}
  {112910} (\bibinfo {year} {2024})}\BibitemShut {NoStop}%
\bibitem [{\citenamefont {Marsden}\ and\ \citenamefont
  {Hoffman}(1993)}]{marsden1993elementary}%
  \BibitemOpen
  \bibfield  {author} {\bibinfo {author} {\bibfnamefont {J.~E.}\ \bibnamefont
  {Marsden}}\ and\ \bibinfo {author} {\bibfnamefont {M.~J.}\ \bibnamefont
  {Hoffman}},\ }\href@noop {} {\emph {\bibinfo {title} {Elementary classical
  analysis}}}\ (\bibinfo  {publisher} {Macmillan},\ \bibinfo {year}
  {1993})\BibitemShut {NoStop}%
\bibitem [{\citenamefont {Sauer}\ \emph {et~al.}(1988)\citenamefont {Sauer},
  \citenamefont {Ahmed-Zaid},\ and\ \citenamefont
  {Kokotovic}}]{sauer1988integral}%
  \BibitemOpen
  \bibfield  {author} {\bibinfo {author} {\bibfnamefont {P.}~\bibnamefont
  {Sauer}}, \bibinfo {author} {\bibfnamefont {S.}~\bibnamefont {Ahmed-Zaid}},\
  and\ \bibinfo {author} {\bibfnamefont {P.}~\bibnamefont {Kokotovic}},\ }\href
  {https://doi.org/10.1109/59.43175} {\bibfield  {journal} {\bibinfo  {journal}
  {IEEE Transactions on Power Systems}\ }\textbf {\bibinfo {volume} {3}},\
  \bibinfo {pages} {17} (\bibinfo {year} {1988})}\BibitemShut {NoStop}%
\bibitem [{\citenamefont {Kokotovic}\ and\ \citenamefont
  {Sauer}(1989)}]{kokotovic1989integral}%
  \BibitemOpen
  \bibfield  {author} {\bibinfo {author} {\bibfnamefont {P.}~\bibnamefont
  {Kokotovic}}\ and\ \bibinfo {author} {\bibfnamefont {P.}~\bibnamefont
  {Sauer}},\ }\href {https://doi.org/10.1109/31.17587} {\bibfield  {journal}
  {\bibinfo  {journal} {IEEE Transactions on Circuits and Systems}\ }\textbf
  {\bibinfo {volume} {36}},\ \bibinfo {pages} {403} (\bibinfo {year}
  {1989})}\BibitemShut {NoStop}%
\bibitem [{\citenamefont {Kokotovic}\ \emph {et~al.}(1976)\citenamefont
  {Kokotovic}, \citenamefont {O'Malley},\ and\ \citenamefont
  {Sannuti}}]{kokotovic1976singular}%
  \BibitemOpen
  \bibfield  {author} {\bibinfo {author} {\bibfnamefont {P.}~\bibnamefont
  {Kokotovic}}, \bibinfo {author} {\bibfnamefont {R.}~\bibnamefont
  {O'Malley}},\ and\ \bibinfo {author} {\bibfnamefont {P.}~\bibnamefont
  {Sannuti}},\ }\href {https://doi.org/10.1016/0005-1098(76)90076-5} {\bibfield
   {journal} {\bibinfo  {journal} {Automatica}\ }\textbf {\bibinfo {volume}
  {12}},\ \bibinfo {pages} {123} (\bibinfo {year} {1976})}\BibitemShut
  {NoStop}%
\bibitem [{\citenamefont {Petzold}(1982)}]{petzold1982differential/algebraic}%
  \BibitemOpen
  \bibfield  {author} {\bibinfo {author} {\bibfnamefont {L.}~\bibnamefont
  {Petzold}},\ }\href {https://doi.org/10.1137/0903023} {\bibfield  {journal}
  {\bibinfo  {journal} {SIAM Journal on Scientific and Statistical Computing}\
  }\textbf {\bibinfo {volume} {3}},\ \bibinfo {pages} {367} (\bibinfo {year}
  {1982})}\BibitemShut {NoStop}%
\bibitem [{\citenamefont {Saccomani}\ \emph {et~al.}(2003)\citenamefont
  {Saccomani}, \citenamefont {Audoly},\ and\ \citenamefont
  {D'Angiò}}]{saccomani2003parameter}%
  \BibitemOpen
  \bibfield  {author} {\bibinfo {author} {\bibfnamefont {M.~P.}\ \bibnamefont
  {Saccomani}}, \bibinfo {author} {\bibfnamefont {S.}~\bibnamefont {Audoly}},\
  and\ \bibinfo {author} {\bibfnamefont {L.}~\bibnamefont {D'Angiò}},\ }\href
  {https://doi.org/10.1016/s0005-1098(02)00302-3} {\bibfield  {journal}
  {\bibinfo  {journal} {Automatica}\ }\textbf {\bibinfo {volume} {39}},\
  \bibinfo {pages} {619} (\bibinfo {year} {2003})}\BibitemShut {NoStop}%
\end{thebibliography}%

\appendix

\section{Additional Results: Manifold Learning}
\label{sec:OutputInformedDiffusionMapsSI}
\begin{figure*}[ht!]
    \centering    
    \includegraphics[width=0.5\linewidth]{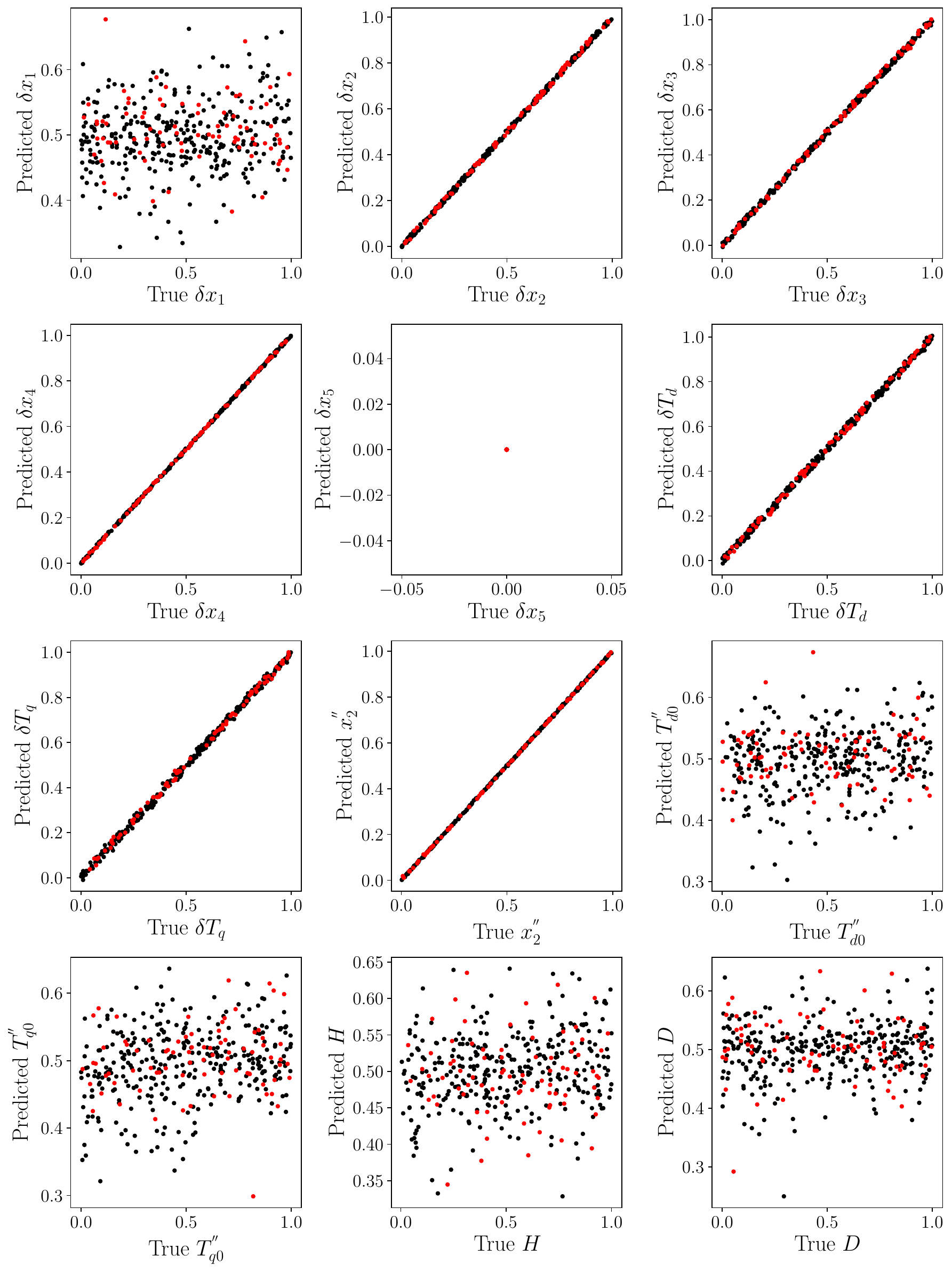}
    \caption{The true (rescaled to 0-1) values of the parameters are plotted against the predicted values using Double DMaps Geometric Harmonics. Black dots represent the training points, while red dots represent the test points.}
\label{fig:parity_plots_GH_parameters}
\end{figure*}

In this section, we present additional complementary results for the Diffusion Maps analysis for the infinite bus system. In the main text, we provided the mean absolute estimates for our regression schemes along with box plots for each output. Here, in Figure \ref{fig:parity_plots_GH_parameters}, we present parity plots comparing the true values of the independent parameters with those estimated using Double DMaps - Geometric Harmonics. Note, that the values have been rescaled  between 0 and 1 for our regression schemes. The black points represent predictions on the training set, while the red points show predictions on the test set. Figure \ref{fig:parity_plots_GH_parameters} also includes the constant independent parameter $\delta x_{5}$ that we omit from the analysis in the main text.

\begin{figure*}[ht!]
\centering
\includegraphics[width=0.5\linewidth]{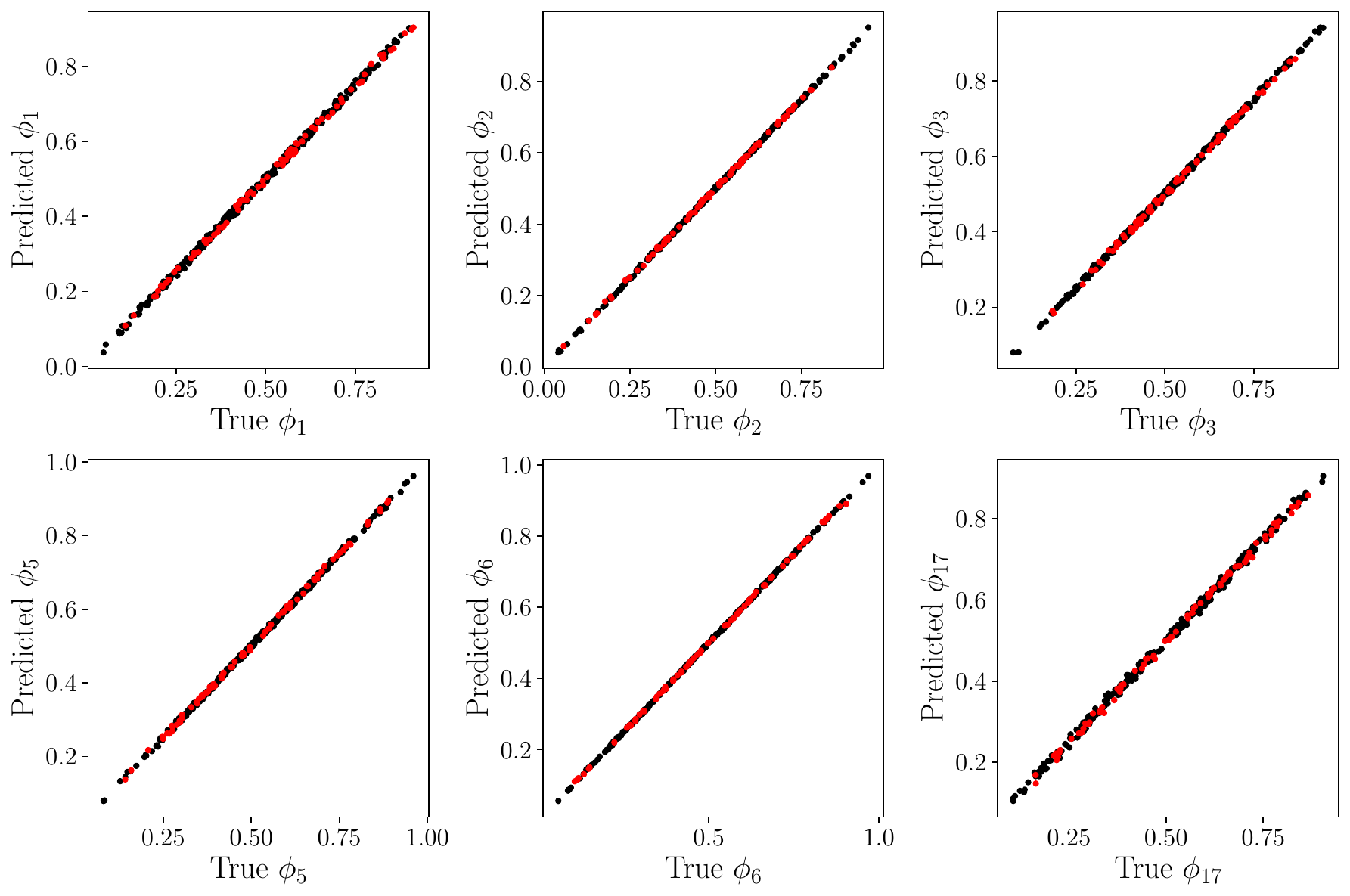}
    \caption{The true (rescaled to 0-1) values of the Diffusion Maps coordiantes are plotted against the predicted values using Double DMaps Geometric Harmonics. Black dots represent the training points, while red dots represent the test points.}
\label{fig:parityplotsDMAPScoordinates}
\end{figure*}

Additionally, in Figure \ref{fig:parityplotsDMAPScoordinates}, we illustrate the parity plots generated based on the inverse mapping $f^{-1}:\vect{p} \mapsto \vect{\phi}$.

\end{document}